\documentclass{article}

\usepackage[preprint]{tmlr}

\usepackage[T1]{fontenc}
\usepackage[utf8]{inputenc}
\usepackage{graphicx}
\usepackage{booktabs}
\usepackage{tabularx}
\usepackage{amsmath,amssymb}
\usepackage{xcolor}
\usepackage[hidelinks]{hyperref}
\usepackage{seqsplit}
\graphicspath{{figures/}}

\newcommand{\evtag}[1]{\allowbreak\textsuperscript{\,\scriptsize\textcolor{black!55}{\texttt{[#1]}}}}

\title{Clean Engineering, Unstable Measurement: A Preregistered Reliability
Failure of Black-Box LLM Observers on Shared Endpoints}

\author{% Real author block for the arXiv preprint / camera-ready builds ONLY.
\name Haoyuan Zhu \email hzhu51@sheffield.ac.uk \\
\addr School of Electronic and Electrical Engineering, University of
Sheffield, Sheffield S10 2TN, UK
\AND
\name Jie Zhang\thanks{Corresponding author.} \email jie.zhang@ranplanwireless.com \\
\addr R\&D Department, Ranplan Wireless Network Design Ltd., Cambridge
CB23 3UY, UK \\
\addr R\&D Department, Cambridge AI+ Ltd., Cambridge CB23 3UY, UK
}

\begin{document}

\maketitle

\begin{abstract}
Language-model judges now gate training data, score generations, and drive leaderboards. The judge is then a measurement instrument, resting on one rarely stated assumption: the same request, sent to the same model name, reads the same tomorrow. We audited that assumption in two preregistered campaigns built to test whether an observer can read solution progress from partial reasoning traces; every threshold was fixed in advance, and neither campaign got past validating its instrument. Across 52,988 audited request attempts, same-window repeat rankings agreed at Spearman 0.400 against a required 0.90, and byte-identical next-day replays agreed at 0.78 against a required 0.99. Those attempts are audit volume, not sample size: the analyses rest on 31 valid task groups, 100 replay pairs, ten windows per supplementary arm, and 3,060 constructed-error judgments. The engineering was not the problem: delivery, schema validity, request hashes, and recorded metadata all sat at ceiling. Three mechanisms explain the gap: a label-to-meaning mapping that biased readouts as strongly as the signal itself; candidate gaps seven orders of magnitude below the instrument's own noise floor; and byte-identical inputs returning different rankings, a noise that exact-permutation readouts compound. Neither metric substitution nor sampling repaired it on the tested grid (a 748,000-call simulated design passed 0 of 500 times). Preregistered follow-ups bound the problem. Waiting did not help on the days sampled (0.805 same-day versus 0.800 cross-day; replicated over five further days, one day's own floor lower still). Switching providers did not help: four providers in three jurisdictions share the floor (medians 0.74 to 0.88), predicted by none of the metadata fields they expose. Self-hosting on batch-invariant kernels helped only while the server was quiet: concurrent load raised disagreement 8.4-fold, back to shared-endpoint magnitude. And the readout is no magnitude meter: on constructed errors with known gaps, its separation tracks error type, not size. We distill the evidence into a three-level snapshot-identity ladder, eight design rules, and a reporting checklist; a pilot at roughly 2\% of the study's call volume would have exposed both unreachable gates in advance. All results concern externally measured behaviour on shared serving infrastructure; they say nothing about model internals and do not assess any provider's service quality. On a shared endpoint, a model name is not a frozen instrument; a preregistered evaluation must measure its instrument before freezing any gate on it.

\end{abstract}

\section{Introduction}

Language models now sit inside the measurement loop of their own field: as judges they rank candidate responses, score reasoning traces, gate training data, and drive leaderboards. Using a model as an instrument carries an assumption so basic it is rarely stated: that the instrument is \emph{stable}. The same request, sent to the same model name, should read the same tomorrow; a frozen evaluation criterion should test the system under study rather than the fluctuating state of its measuring device. On shared inference endpoints this assumption is not innocent. The systems literature documents why: serving infrastructure batches concurrent requests, common inference kernels are not batch-invariant, and deployments change behind fixed model names (\S{}2); and none of this is visible in the response metadata an evaluator can record, so a black-box study can observe the resulting instability without being able to attribute it. The question this paper answers is the metrological one that the judge paradigm has mostly skipped: \textbf{when an LLM observer is held to the standard of a scientific instrument (preregistered thresholds, frozen protocols, no post-hoc repair), does it pass? And when it does not, can the platform's own metadata even reveal why?}

We did not choose this question; we collided with it. A preregistered research programme (measuring correctness-probability paths over visible reasoning prefixes) required its observer to pass instrument-validation gates before any scientific claim could be tested. It never got further. Across two campaigns and 52,988 audited request attempts \evtag{index@cf4ea148 \S{}8}, both terminated at the instrument layer: a same-window repeat-ranking gate failed at Spearman 0.400 against a frozen 0.90, and a next-day byte-identical replay gate failed at 0.78 agreement against a frozen 0.99, each time with the execution record at ceiling. Every request delivered, every schema valid, every request body hash-identical to its frozen plan, every recorded metadata field constant. The engineering record showed nothing wrong; the preregistered criteria declared the instrument unusable. This paper is the full audit of that contradiction, and of what it teaches about using black-box observers on shared endpoints.

Four literatures surround the question, surveyed in \S{}2: judge reliability (biases catalogued early; self-consistency recently measured \citep{zheng2023mtbench,roulette2025,irtjudge2026}), the nondeterminism mechanism (batch-dependent kernels, batch-invariant remedies \citep{tml2025nondeterminism,vllm-batch-invariance}), reproducibility machinery (checklists and preregistration proposals \citep{pineau2021repro,reforms2024,miltenburg2021prereg}), and a measurement-theoretic turn \citep{jacobs2021measurement}. Each stops short of the instrument-audit setting. Self-consistency is reported descriptively: no study we found lets the number act as a preregistered gate with authority to terminate a research line, ties it to a byte-identity audit chain, or decomposes it into same-day floor versus cross-day change. The nondeterminism literature explains the mechanism but does not quantify it for structured ranking readouts under frozen criteria, where one token flip voids a whole record. The checklists require versions and seeds but no snapshot-identity level and no measured metadata semantics, while vendors themselves document that a fixed model ID does not fix observable behaviour \citep{anthropic-model-ids}. And the \emph{design-time} discipline of analytical chemistry and industrial inspection (measure the noise floor and the gap distribution first, run the criterion's operating characteristic, only then freeze a threshold \citep{currie1999lod,aiag2010msa,iso2859}) has, as far as our retrieval found, never been applied to an LLM evaluation gate. Our own campaigns sat inside that last gap, which is precisely how a threshold comes to test nothing at full ceremony.

We turn the failed programme into the audit the field lacks, in four moves: publish the failure whole, under an append-only audit chain with every excluded window disclosed and every number resolving to a SHA-256-pinned archive member (\S{}3--\S{}4, \S{}12); dissect the mechanism into three modes that one permutation readout amplifies into a single observable (\S{}5--\S{}7); attribute the instability prospectively, with preregistered same-day/cross-day, four-provider, self-hosted, and constructed-error supplements (\S{}8, \S{}10); and convert the evidence into reusable discipline (\S{}9, Appendix D).

The ingredient facts are not new: judge inconsistency, temperature-zero nondeterminism and its kernel-level mechanism, position bias, and reporting checklists are all established (\S{}2, Table 1). What this paper adds is the combination none of that work executed: a fully preregistered \emph{audit} of the stability assumption with frozen thresholds and no post-hoc repair, the mechanism decomposition of its failure, prospective attribution and calibration supplements, and the instrument-first discipline the failure licenses. The claims are scoped to the tested configurations throughout; the transferable product is the discipline, not a universal law.

\textbf{Contributions.}

\begin{enumerate}
\item \textbf{A fully audited preregistered negative result} (\S{}3--\S{}4, \S{}12): two independent instrument generations failing their frozen gates with execution records at ceiling, establishing with unusual evidential hygiene that engineering-layer perfection carries no implication about measurement-layer reliability, and that the two must be reported separately.
\item \textbf{A mechanism decomposition of judge instability} (\S{}5--\S{}7): instrument bias, near-degenerate score separations, and platform nondeterminism, unified by the amplifier property of permutation readouts; the two natural escapes (metric substitution and sample-size scaling) closed on the tested grid; and the resolution question checked from outside the instrument's own loop with a constructed-error battery (\S{}10.3).
\item \textbf{Attribution experiments} (\S{}10): a preregistered same-day/cross-day variance decomposition of judge replay stability with dependence-aware uncertainty, a four-provider characterisation showing the floor at comparable magnitude on every tested platform, and a direct measurement of whether the \texttt{system\_fingerprint} field predicts stability (it does not, in three distinct failure modes).
\item \textbf{The audit boundary, stated as levels} (\S{}8): the L0--L2 snapshot-identity ladder; a taxonomy of boundary forms found by instrumenting the audit itself (documentation contradicted by measured behaviour in four instances, a protocol-level ban on deterministic settings, bimodal rate-limit observability, aliases as moving targets); and the reporting implication that snapshot identity belongs beside temperature and seed as mandatory metadata.
\item \textbf{An instrument-first evaluation discipline} (\S{}9, Appendix D): the pilot--simulate--freeze pipeline with operating-characteristic calibration, priced by this study's own history at pilot scale (about a thousand calls against the fifty-three-thousand-attempt sequence it would have redirected); and the reporting checklist extending existing community checklists with the instrument-status items they lack.
\end{enumerate}

\textbf{What this paper does not claim.} None of the results here bear on the internal states or semantic dynamics of the observed models: every quantity is an external measurement over visible records, and the theory-testing programme that motivated the instrument lives in a separate white-box track with its own preregistration. Nor is any result a quality assessment of a named provider: the behaviours we document are consistent with the public engineering properties of shared inference services, and our task family manufactured its own worst case (\S{}11.2). The claim is narrower and, we believe, more useful: \textbf{on a shared endpoint, a model name is not a frozen instrument, and a preregistered evaluation must measure its instrument before freezing any gate on it.} Sections 3--8 establish that sentence empirically; sections 9--13 turn it into practice.

\section{Background and related work}

\textbf{Judge reliability and nondeterminism.} The judge paradigm entered wide use with MT-Bench and Chatbot Arena \citep{zheng2023mtbench}, and its systematic defects were catalogued early: position bias strong enough that reordering two candidates flips a verdict \citep{wang2024fair,shi2024positionbias}, verbosity and self-enhancement biases \citep{zheng2023mtbench}, self-preference measurable at the mechanism level \citep{panickssery2024selfpref}; surveys organise the space \citep{gu2024survey,li2024judgessurvey}, with agreement with human judgment, measured once, as the dominant criterion \citep{bavaresco2024judgeval}. A recent line measures the judge's agreement \emph{with itself}: repeated-call self-consistency is generally low across NLG tasks \citep{roulette2025}, large-scale audits separate reliability from validity \citep{rwv2026}, and item-response theory has been applied to judge reliability \citep{irtjudge2026}. Those studies characterise reliability descriptively across repeated runs, typically jointly with human agreement or fidelity, and their numbers inform but do not bind: no reported consistency value carries authority over the study that measures it. Here the same quantity is a preregistered gate with the authority to terminate a research programme, every request is byte-auditable, and the instrument is calibrated before interpretation; that enforcement difference, not the measurement of inconsistency, is the contribution. The instability's mechanism is settled engineering knowledge: shared endpoints batch concurrent requests, kernels are not batch-size-invariant, and greedy decoding atop drifting logits diverges, with batch-invariant kernels restoring bitwise equality at known cost \citep{tml2025nondeterminism,vllm-batch-invariance}. Empirically, instability at nominally deterministic settings is documented for code generation \citep{ouyang2025nondeterminism} and task suites \citep{atil2024nondeterminism}, contaminates security evaluations \citep{pretender2026}, and cross-\emph{version} drift is established for dated snapshots months apart \citep{chen2024behavior}; prompt-sensitivity work varies the input and measures spread \citep{sclar2024formatspread}, while our replays hold the input fixed to the byte, the orthogonal axis. We claim no novelty for the phenomenon. Our contribution to both lines is the \emph{instrument-audit} setting (Table 1): preregistered thresholds that a research programme enforced, byte-level identity that makes the instability attributable, its decomposition into bias, separation, and platform terms, and frozen-band quantification for structured ranking readouts (\S{}5, \S{}10). M1 is the readout-level face of position bias, adding the interaction and ranking cascade that literature rarely quantifies (\S{}5.1); and where \citep{chen2024behavior} compares \emph{different} dated snapshots over months, we show exact replay failing \emph{within} one dated snapshot on one afternoon (0.805), bracketing the audit problem from the opposite side.

\textbf{Reproducibility machinery and measurement science.} Community checklists (NeurIPS's \citep{pineau2021repro}, NLP's \citep{magnusson2023checklist}, REFORMS \citep{reforms2024}) require versions, hyperparameters, code, and compute; none requires what \S{}8 shows is load-bearing for API-hosted evaluation: a snapshot-identity level, the response-side model distribution, or measured metadata semantics. Harness builders document API-side irreproducibility from experience \citep{biderman2024lessons}, vendors document that fixed model IDs do not fix observable behaviour \citep{anthropic-model-ids}, and the nearest governance work gates model updates on the supply side \citep{supplychain2026}; our checklist (\S{}9.7, Appendix D) extends these with the instrument-status items they lack. \citep{miller2024errorbars} supplies report-time uncertainty; our gate calibration (\S{}9.2) is the design-time counterpart. Preregistration has been proposed and debated for NLP \citep{miltenburg2021prereg,sogaard2023prereg} but remains rare in executed form; this paper is among the few cases where preregistered gates were enforced to a terminal negative verdict with the full audit trail published (\S{}11.5). The questions themselves were formalised long ago in analytical chemistry and industrial quality control: detection limits \citep{currie1999lod,iupac-goldbook-lod}, gauge R\&R \citep{aiag2010msa}, operating characteristics in acceptance sampling \citep{iso2859}, equivalence bands \citep{lakens2017tost}. ML's measurement-theoretic turn \citep{jacobs2021measurement,benchepistemology2025,irtjudge2026} concentrates on construct validity and report-time variance; the \emph{instrument-first} discipline (noise floor and gap distribution first, operating characteristic next, threshold last) has, as far as our retrieval found, no precedent in LLM evaluation. Sections 5, 7, and 9 import it whole, with this study's two campaign failures as the price of learning it the other way around.

\textbf{Positioning.} This paper contributes no new judging method and no new benchmark. It contributes a fully audited negative result, the mechanism decomposition that explains it, prospective supplements that attribute it, and the design rules and checklist the evidence supports. Its nearest neighbours each hold one piece: self-consistency measurement without preregistered enforcement, nondeterminism mechanisms without readout-level audit, update governance without downstream reporting levels, measurement theory without gate design. The claim to attention is the assembled whole, checked against its own archive at every number; Table 1 states the division line by line.

\textbf{Table 1: Contributions relative to prior work.}

\begin{center}
\small
\begin{tabularx}{\textwidth}{>{\raggedright\arraybackslash}X>{\raggedright\arraybackslash}X}
\toprule
Established before this paper & This paper adds \\
\midrule
Judge biases: position, verbosity, self-preference \citep{zheng2023mtbench,wang2024fair,panickssery2024selfpref} & a bias diagnostic \emph{inside} the readout (\(b_{\mathrm{map}}\)), measured against the signal it contaminates, and the symmetrisation that cancels its first order (\S{}5.1) \\
Judges' self-agreement across repeated calls, reported descriptively alongside human agreement or fidelity \citep{roulette2025,rwv2026,irtjudge2026} & the same quantity as an \emph{enforced} preregistered gate (frozen threshold, no-repair rule, authority to terminate), byte-auditable, and decomposed into bias, separation, and platform terms (\S{}4--\S{}5) \\
Temperature-zero drift traced to batch-dependent kernels; batch-invariant inference as remedy \citep{tml2025nondeterminism,vllm-batch-invariance} & what the documented mechanism does to a \emph{preregistered evaluation}: the permutation-readout amplifier, and same-day/cross-day plus four-provider floors under frozen bands (\S{}6, \S{}10) \\
Behavioural drift across dated snapshots months apart \citep{chen2024behavior} & drift under \emph{byte-identical requests within one dated snapshot name}, with every recorded metadata field constant (\S{}5.3, \S{}8) \\
Reporting checklists and preregistration proposals \citep{pineau2021repro,reforms2024,miltenburg2021prereg,biderman2024lessons} & the instrument-status items they lack (noise floor, gap distribution, fingerprint status, snapshot-identity level), each traced to a measured failure here (\S{}9, App. D) \\
Judges and benchmarks framed as measurement instruments \citep{jacobs2021measurement,benchepistemology2025,irtjudge2026} & an \emph{executed} instrument-validation campaign under that framing: gates, calibration, detection limits, audit boundary, and the full negative result disclosed (\S{}3--\S{}8, \S{}12) \\
\bottomrule
\end{tabularx}
\end{center}

\section{Measurement setup, preregistration discipline, and audit chain}

The failures reported in this paper are only informative if the instrument that produced them was specified before execution, executed exactly as specified, and audited byte-by-byte afterwards. This section describes that instrument. All design constants below were frozen in configuration snapshots and preregistered gate definitions before any paid call was placed; the audit chain that enforces this is described in \S{}3.5. Figure~\ref{fig:pipeline} shows the measurement chain end to end, and the terminology map in Appendix A.12 collects the paper's recurring names so no reader needs the project history to follow the argument.

\subsection{Measured object and operationalisation}

The measured system is a two-model pipeline over deterministic arithmetic tasks. A \emph{generator} produces, for each task, four candidate solutions written as explicit step-by-step traces (4--8 visible steps) ending in one exact rational answer. Each task has a unique canonical answer, and the only supervision label in the entire study is exact correctness of a candidate's final answer, \(Y_{qn}\in\{0,1\}\), decided by canonical rational comparison, never by a model. An external \emph{observer} is then shown the task together with a visible prefix of one candidate's trace, truncated at progress level \(j\in\{1,\dots,4\}\) (25\%, 50\%, 75\%, 100\% of the visible steps), and is read out as a Bernoulli judgement \(p_{qnj} = P_{\mathrm{obs}}(Y_{qn}=1 \mid \text{task},\ \text{visible prefix})\).

Two declarations bound everything that follows. First, every quantity in this paper is an operationalisation of \emph{an external measurement applied to a visible record}; nothing observes, or is claimed to observe, an internal state of the generator (\S{}11.3). Second, the study never reached its scientific objective: both preregistered campaigns terminated at the instrument-validation layer, because the gates below failed. The data analysed here are therefore instrument data, and are used for nothing else.

\subsection{Three readout protocols (two instrument generations)}

\begin{center}
\small
\begin{tabularx}{\textwidth}{l>{\raggedright\arraybackslash}X>{\raggedright\arraybackslash}X}
\toprule
Generation & Protocol & Readout \\
\midrule
Legacy D1 & Single-mapping exact two-label logprob: the observer must emit exactly \texttt{A} or \texttt{B}; both exact labels are read at the single answer-token position & \(r = e^{\ell_+}/(e^{\ell_+}+e^{\ell_-})\) \\
D1-S (v2.3a) & Dual-mapping adjacent pairs: each measurement point is asked twice, with the label-to-meaning mapping reversed between arms; probabilities are clipped at \(\varepsilon=10^{-4}\) before the logit & \(z_{\mathrm{sym}}=(z^{(+)}+z^{(-)})/2\), with mapping-bias diagnostic \(b_{\mathrm{map}}=(z^{(+)}-z^{(-)})/2\) \\
I0-R judge & Structured full-ranking output: one JSON object whose \texttt{ranking} field is an exact permutation of ten presented candidate aliases & the permutation itself \\
\bottomrule
\end{tabularx}
\end{center}

The symmetrised protocol exists because the legacy protocol failed its stability gate in a way traceable to label semantics (\S{}5.1); the judge protocol exists because the second campaign selected it as primary observer protocol on frozen development data (\S{}3.4). The progression matters for what follows: each generation \emph{repaired the previous generation's diagnosed failure mode}, and each then failed on a different preregistered criterion.

\subsection{Frozen configuration}

Every design constant was sealed in a run-local, hash-pinned configuration snapshot before execution; the full frozen configuration of the I0-S run (task design, generator, observer, protocol, request plan, budget) and the judge-protocol campaigns' shared wire body are reproduced in Appendix A.9. Three constants matter for what follows: the observer ran at temperature 0.0 with one-token exact-label readouts (\texttt{top\_logprobs} 5, thinking disabled) over 80 tasks $\times$ 4 candidates, on a fixed plan of 3,312 calls under a hard cap of 3,632. The cap is enforced by a counter that reserves \emph{before} each network attempt and counts retries and transport failures, so a run cannot exceed it by construction. Arm order within pairs was balanced by frozen seed: of the 1,496 pairs, 748 ran forward-first and 748 reverse-first \evtag{phase1-i0s-fresh@ac06bb97}.

\subsection{Preregistered gates}

Each campaign froze a battery of instrument gates before execution; the full definitions are reproduced in Appendix A. The two that decide this paper's narrative:

\textbf{Gate I-3S (instrument stability, I0-S).} Five subcriteria, all frozen: single-arm exact two-label coverage $\geq$ 0.995; pair completeness $\geq$ 0.990; arm label mass median $\geq$ 0.95 with 1st percentile $\geq$ 0.80; repeated-pair scalar stability, median \(|\Delta r_{\mathrm{sym}}|\le 0.03\); and repeated candidate-ranking stability, within-task Spearman median $\geq$ 0.90 over the 10\% repeated subset.

\textbf{Gate I-5R (next-day replay, I0-R).} After a mandatory $\geq$ 24 h gate, 100 preregistered judge requests are replayed with byte-identical request bodies against the same model name; the frozen criterion is full-ranking agreement (exact permutation match; recorded as "structured-field agreement" in the frozen artifacts) $\geq$ 0.99, with every preregistered canary kept in the denominator.

Missing labels invalidate the affected arm and pair; no tail-mass allocation, no imputation, and no post-hoc metric substitution are permitted anywhere in the battery \evtag{phase1-i0s-fresh@ac06bb97}. One preregistration deviation in the second campaign is disclosed and bounded in \S{}11.5: the calibration-stage freeze postdates two development results, so this paper never describes I0-R as fully pre-frozen.

\subsection{Audit chain}

Every network attempt appends, before and after the call, to an append-only JSONL event log inside the run directory: request body with canonical and wire-order SHA-256s, raw response with its SHA-256, UTC timestamps, verbatim provider metadata, retry count (the full field reference is Appendix C). Interrupted runs may resume only if the existing log is a strict prefix of the frozen plan. On completion a manifest records the SHA-256 of every artifact, the directory becomes immutable, and verification tooling re-hashes every sealed run, failing on any drift. The I0-S full run and the F2 replay window are anchored by manifests \evtag{phase1-i0s-fresh@207cd6eb} and \evtag{i0r-round@22c05608}.

\begin{figure}[t]
  \centering
  \includegraphics[width=\linewidth]{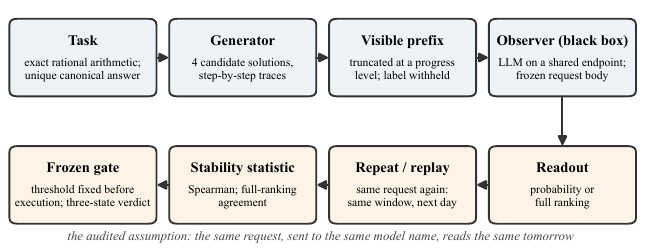}
  \caption{The measurement chain. A generator writes four candidates per
  exact-arithmetic task; a black-box LLM observer sees the task plus a
  truncated trace prefix and returns a probability or full ranking;
  repeats and replays feed a stability statistic against a threshold fixed
  before execution. The paper audits the italicised assumption.}
  \label{fig:pipeline}
\end{figure}

\section{Engineering correctness does not imply measurement reliability (C1)}

Both preregistered campaigns produced the same shape of event: an execution record with no defect anywhere a software engineer would look, and a scientific gate that failed exactly on its frozen threshold. This section establishes that shape as a fact about the instrument rather than about our code, because every downstream claim (C2--C6) interprets it.

\subsection{Two clean failures, side by side}

\textbf{Table 2: Execution integrity versus preregistered verdict.} Left column: the I0-S full run (same-window repeat stability). Right column: the I0-R Stage 3-F2 window (next-day exact replay). Sources: \evtag{phase1-i0s-fresh@ac06bb97}, \evtag{i0r-round@55674b09}, \evtag{i0r-round@e8420c2f}.

\begin{center}
\small
\begin{tabularx}{\textwidth}{>{\raggedright\arraybackslash}X>{\raggedright\arraybackslash}X>{\raggedright\arraybackslash}X}
\toprule
 & I0-S (same-window repeat) & I0-R Stage 3-F2 (next-day replay) \\
\midrule
Planned calls & 3,312 fixed (cap 3,632) & 100 fixed (cap 200) \\
Calls with a provider response & 3,312 / 3,312 & 100 / 100 \\
Retries / transport errors / refusals & 0 / 0 / 0 & 0 / 0 / 0 \\
Schema-valid responses & pairs stored 1,496 / 1,496 & 100 / 100 \\
Request bodies vs frozen plan & plan enforced by construction & SHA-256 match 100 / 100 \\
Response \texttt{model} field & constant across run & identical 100 / 100 \\
\texttt{system\_fingerprint} & --- & identical 100 / 100 (both windows: \texttt{null}) \\
Finish reason & uniform & identical 100 / 100 \\
Exact-label coverage / pair completeness & 1.0 / 1.0 & n/a (structured branch) \\
Arm label mass, median (p01) & 0.999475 (0.997172) & n/a \\
\textbf{Scalar repeat stability} & median \(|\Delta r_{\mathrm{sym}}|\) = \textbf{0.006710} (threshold $\leq$ 0.03: passes with 4.5$\times$ margin) & --- \\
\textbf{Preregistered primary verdict} & ranking Spearman median \textbf{0.400} < 0.90 $\rightarrow$ \textbf{Gate I-3S fail} & full-ranking agreement \textbf{78/100 = 0.78} < 0.99 $\rightarrow$ \textbf{Gate I-5R fail} \\
\bottomrule
\end{tabularx}
\end{center}

Read each column top to bottom. Everything an execution audit can check (delivery, schema validity, byte-identity of requests, constancy of the recorded metadata set: response \texttt{model} string, \texttt{system\_fingerprint}, finish reason, HTTP status) is at ceiling in these two runs. The preregistered scientific criterion then fails, and not marginally: I-3S fails at 0.400 against a 0.90 threshold while the \emph{scalar} readout of the same instrument repeats to within 0.0067; I-5R fails at 0.78 against 0.99 while each of the recorded metadata fields reports that nothing changed. The engineering context around the table is equally clean: 209/209 offline tests passing at the failing commits, F2 manifest artifacts verified 11/11, zero credential-shaped fields in the audit logs, no batching API, zero retries across the I0-S plan \evtag{docs-snapshot@7440c20c}. Appendix C lists the complete recorded field set; "metadata" in this paper always means that set, not an unbounded claim about everything a platform could expose.

Two further properties make the pair of failures evidential rather than anecdotal. They come from \emph{different instrument generations} (a symmetrised two-label logprob readout; a structured full-ranking judge) that were designed two days apart, each repairing the previous generation's diagnosed defect. And they fail on \emph{different axes} (repeat stability within a window; exact replay across a 24-hour gate) that share no statistic.

\subsection{The failures were executed faithfully}

A failed gate is only worth publishing if the failure was not managed away. The frozen protocols left several tempting exits, and the record shows none was taken \evtag{docs-snapshot@a021936f}:

\begin{itemize}
\item thresholds were not lowered, and the primary metric was not switched after the fact (the tempting substitutions are quantified in \S{}6: the same F2 drift reads 95/100 under top-1 agreement and 4436/4500 $\approx$ 0.986 under pairwise agreement \evtag{i0r-round@3be62ed8}, and citing either as the verdict was preregistered out);
\item all 22 drifted replay records stay in the denominator; nothing was deleted, re-run, or replay-selected, and no post-hoc majority vote was taken;
\item the two candidates whose final answers failed to parse are retained as failures (318/320 parseable), not imputed;
\item the fallback observer protocol (verbal confidence) was not activated: a scientific gate failure does not trigger it;
\item the I-3S failure is broad, not an outlier artifact: 29 of 31 valid rank groups sit below the threshold, and every prefix progress level has a median below 0.90 \evtag{docs-snapshot@7440c20c}.
\end{itemize}

The same discipline extends backwards into the record: every abandoned window, blocked endpoint, schema-defect batch, and rejected design in either campaign is retained and disclosed (\S{}12.1), and none of them enters any denominator in this paper.

\subsection{The claim}

\begin{quote}
\textbf{C1.} On a shared endpoint, full marks on transport-level and engineering-level compliance carry \emph{no implication} about measurement-level reliability. "All tests green and all hashes match" is a necessary condition for a preregistered evaluation, and nothing more. Reports must therefore state the two layers separately, and a reproducibility claim that cites only execution-layer evidence (delivery rates, schema validity, byte-identical requests, constant metadata) has asserted nothing about whether the instrument would read the same tomorrow.
\end{quote}

The remainder of the paper dissects the gap that C1 exposes: \S{}5 locates the mechanisms (a near-degenerate measurand; a platform that does not promise determinism; gates frozen on never-measured quantities), \S{}6--\S{}7 close the two natural escapes (metric substitution, more sampling), \S{}8 states what the platform's own metadata can and cannot certify, and \S{}9 turns each mechanism into a design rule.

\begin{figure}[t]
  \centering
  \includegraphics[width=\linewidth]{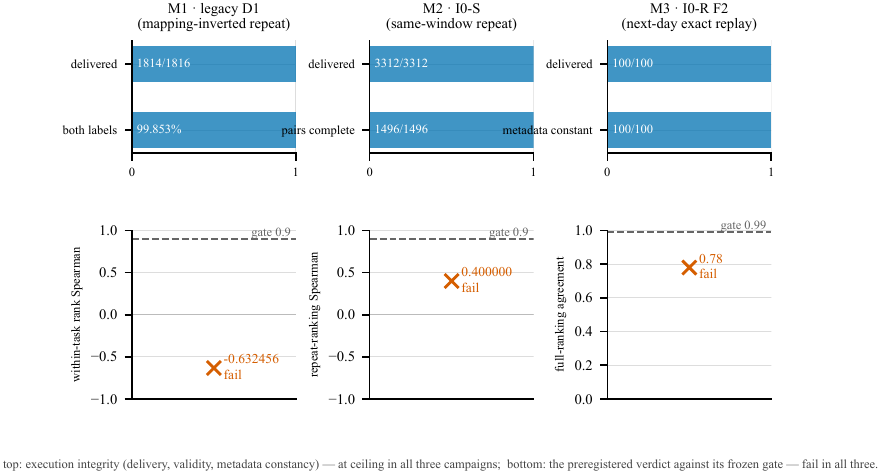}
  \caption{Three failure modes, one shape. Top: execution-integrity rates
  for D1, I0-S, and F2, at or near ceiling (legacy delivery 1,814/1,816;
  label coverage 99.853\%; I0-S and F2 at 100\%). Bottom: each campaign's
  preregistered verdict against its frozen threshold (dashed), all
  failing. The contrast between the rows is claim C1.}
  \label{fig:failure-modes}
\end{figure}

\section{Three failure modes, one root-cause structure (C2)}

Section 4 established that the failures are properties of the measurement. This section characterises them. We name the modes M1--M3 in the order they were encountered; each was diagnosed with the audit data of the run that produced it, and each diagnosis then shaped the next instrument generation.

\subsection{M1: label-semantics mapping bias (legacy D1)}

\textbf{Phenomenon.} The legacy single-mapping protocol passed its scalar repeatability criterion and failed its ranking criterion \emph{with a negative median}: across inverted-mapping repeats, the within-task candidate-ranking Spearman median was \textbf{-0.632} (frozen threshold $\geq$ 0.90), while the median absolute probability change was 0.0233, inside its $\leq$ 0.03 bound \evtag{phase1-oldD1@2dc38b05}. A negative median is the signature of something stronger than noise: flipping which label means "correct" tended to \emph{invert the ranking}.

\textbf{The mundane explanation is excluded.} This is not a token-emission defect: all 1,494 successful responses emitted exactly \texttt{A} or \texttt{B} at one answer position with both exact labels in the top-logprobs (rates 1.0), labels balanced (A: 734, B: 760), model string constant \evtag{docs-snapshot@a8c3e9bb}. The instrument delivered exactly the readout it was asked for; the readout itself was biased.

\textbf{Mechanism.} Define, over an adjacent mapping pair, the symmetrised logit \(z_{\mathrm{sym}}=(z^{(+)}+z^{(-)})/2\) and the mapping-bias diagnostic \(b_{\mathrm{map}}=(z^{(+)}-z^{(-)})/2\). If the readout carried only a correctness signal, \(b_{\mathrm{map}}\) would be zero up to noise. On the 136 legacy inverted pairs it is not: median 0.250 logits, quartiles -0.187 to 0.875, 95th percentile of \(|b_{\mathrm{map}}|\) at \textbf{1.391} logits: the magnitude of the \emph{bias} component rivals the dynamic range of the \emph{signal}. Worse, it is not a constant offset that ranking could ignore: its median is \textbf{+0.531} on correct candidates and \textbf{-0.250} on incorrect ones, i.e. the bias interacts with the quantity being measured. Consistently, rankings computed from forward arms agree with rankings computed from reverse arms at a Spearman median of only 0.211 \evtag{docs-snapshot@33dda04b}. A single-mapping readout therefore confounds label-direction preference with correctness; whichever mapping a protocol happens to fix becomes part of its measurand. The successor protocol (D1-S) cancels the first-order antisymmetric component by construction, with the caveat, carried through \S{}11, that symmetrisation does not guarantee the vanishing of higher-order interactions between label tokens and prompt content. The bias itself persists undiminished in the fresh window: over the 1,496 I0-S pairs, \(b_{\mathrm{map}}\) has median 0.3125, quartiles -0.1875 to 0.7500, and 95th percentile of \(|b_{\mathrm{map}}|\) at 1.3125 logits (Figure~\ref{fig:bmap}; recomputed from the pair data at figure-build time, \evtag{phase1-i0s-fresh@bbed8c1b}). What symmetrisation changed is that the bias no longer \emph{enters the readout}, not that the observer stopped having it.

\subsection{M2: repeat-ranking instability on a near-degenerate measurand (I0-S)}

\textbf{This is the paper's central mechanism.} The symmetrised instrument produced what looks like a paradox: scalar repeatability of median \(|\Delta r_{\mathrm{sym}}|=0.00671\) (4.5 times better than its frozen bound) and a repeat-ranking Spearman median of 0.400 against a 0.90 gate, with 29 of 31 valid groups below threshold. The denominator is stated exactly: of the 32 repeated rank groups, one is constant on repeat and has no defined Spearman; the frozen validity rule excludes it, never imputes it, and ties within valid groups receive average ranks \evtag{docs-snapshot@e975a840}. The resolution is in two distributions measured by the frozen failure audit \evtag{docs-snapshot@57f3389b}:

\emph{What the instrument moves.} Repeat deltas: quartiles 0.001--0.018, 90th/95th percentiles 0.052/0.080, maximum 0.167. Call this the instrument's noise floor: roughly \(10^{-3}\)--\(10^{-2}\) in probability.

\emph{What there is to measure.} Within a task, the four candidates' readouts span a median range of 0.0114; and the median \emph{minimum} gap between adjacent candidates is \(3.04\times10^{-10}\), with the 75th percentile still at 0.0005. Twelve of 32 groups contain exact ties on the original readout, 7 of 32 on the repeat; only 62 of 128 candidates keep even their within-pair arm order across repeats.

A ranking gate asks the instrument to resolve \emph{gaps}, and here the typical gap sits four to seven orders of magnitude below the noise floor. Under repeat, candidates separated by \(10^{-10}\) reorder with probability essentially \(\tfrac12\) each; "scalar stable, ranking collapsed" is not a contradiction but arithmetic. The progress-stratified medians confirm that the instrument detects order exactly where order exists to detect: Spearman medians by visible-prefix level are 0.000 (25\% visible), 0.769 (50\%), 0.500 (75\%), 0.000 (100\%). At one quarter of the trace, candidates have not yet diverged; at completion, the final answer is visible to the observer, readouts saturate, and same-correctness candidates become indistinguishable again; the ranking information lives mid-trace.

\textbf{Upstream cause.} The degeneracy was manufactured by the task design, not discovered in nature: candidate accuracy was 0.781, and only 16.25\% of tasks had candidates of mixed correctness \evtag{docs-snapshot@7b8f55c0}. In roughly 84\% of groups there was \emph{nothing to rank}: all four candidates share a correctness value toward which their probability paths converge. The gate, however, was frozen at 0.90 before any of these intervals had been measured. The design rule this yields (\S{}9.3, \S{}9.6) is the detection-limit discipline of analytical chemistry: a ranking gate is only meaningful relative to the measured gap distribution of its measurand, and the gap distribution must be piloted before the gate is frozen.

One caveat is owed here and repaid prospectively: both distributions come from the same observer, so on the legacy data alone this section characterises \emph{observed score separations}. The controlled-error battery of \S{}10.3 closes that loop from outside, with construction-known errors: the readout separates wrong from right at every magnitude tested, but by error \emph{type}, not error \emph{magnitude}. The instrument discriminates; it does not meter, and a readout that never orders by magnitude cannot certify order among near-degenerate alternatives, whatever its scalar repeatability.

\subsection{M3: byte-identical inputs, drifting outputs (I0-R)}

\textbf{Phenomenon.} The second campaign's judge instrument, replaying 100 frozen requests after a mandatory 24-hour gate with byte-identical bodies (SHA-256-verified 100/100), reproduced the exact ranking 78 times out of 100. The 22 drifted permutations spread across 16 of 25 tasks (not one failing request) while every provider-visible metadata field was constant (\S{}4.1). The drift is \emph{local}: the Kendall-distance spectrum over replays is 78 exact, 11 at distance 1, and 11 spread over distances 2--11; top-1 agreement is 95/100 and pairwise order agreement 4436/4500 $\approx$ 0.986 \evtag{i0r-round@3be62ed8}. Mostly, these are single adjacent transpositions: exactly what near-degenerate subsets (M2) plus small numerical perturbations produce in a full-permutation readout.

\textbf{What the drift is not.} We originally named this mode "overnight drift", implying attribution to backend change across the 24-hour gate. Our prospective supplement A-S1 (\S{}10.1) falsifies that reading: replaying the same battery five times \emph{within} each of two days and five times across them gives a same-day full-agreement median of 0.805 and a cross-day median of 0.800 (difference 0.005, inside the preregistered equivalence band), with the historical 0.78 sitting inside the same-day distribution itself \evtag{index@cf4ea148 appendix A.3}\evtag{repo:docs/reports/a\_s1\_samedayvar\_r1.md}. The correct statement of M3 is: \textbf{on this platform class, exact input identity does not purchase exact output identity even within an afternoon; the 24 hours added nothing}. The known mechanism, batch-size-dependent kernel scheduling under shared load (\S{}2.2), is consistent with everything we observe, and A-S2 (\S{}10.2) shows the same floor on four providers.

\subsection{The shared structure}

\begin{quote}
\textbf{Root cause 1 (measurand).} The score separations to be resolved are near zero in the readout's own units, and the readout is in any case not ordered by error magnitude (M2, \S{}10.3). \textbf{Root cause 2 (platform).} The tested shared endpoints do not promise bitwise determinism at temperature 0 and expose no auditable snapshot identity; a nonzero noise floor persisted across every window we measured, and nothing recordable at L0 could attribute or remove it (M3, \S{}8). \textbf{Root cause 3 (process).} Both gates were frozen on quantities that had never been pilot-measured (a repeat-ranking statistic, I-3S, and an exact replay-agreement rate, I-5R), so neither threshold was reachable by any instrument of the class being audited (\S{}9.2).
\end{quote}

M1 is a biased instrument, M2 a degenerate measurand, M3 an unstable platform: three distinct defects that produce the \emph{same observable} in ranking readouts, because a permutation is an amplifier: an infinitesimal scalar perturbation becomes a discrete, gate-visible rank event whenever it crosses one of many near-zero gaps. This is why \S{}9.4 recommends continuous readouts with equivalence bands over exact-match permutation gates, and why \S{}6 shows that no metric substitution repairs the underlying degeneracy.

\begin{figure}[t]
  \centering
  \includegraphics[width=\linewidth]{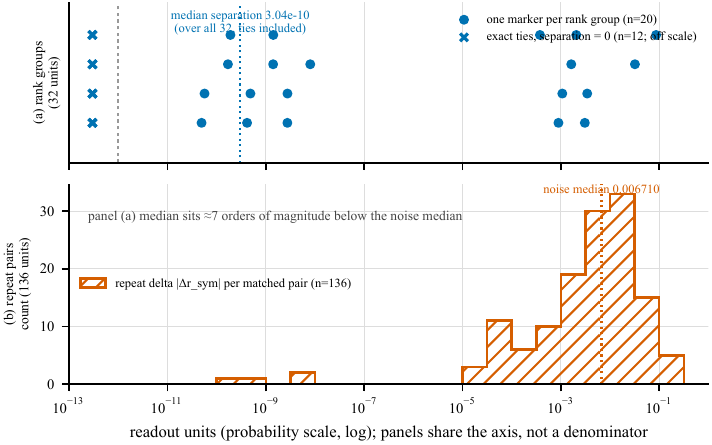}
  \caption{The detection-limit picture behind M2, per sampling unit.
  (a) The 32 repeated groups' minimum score separations, the 12 exact-tie
  groups as a zero-gap category a log axis cannot show. (b) The 136
  repeat-pair deltas, the noise floor. Panels share the axis, not a
  denominator; the medians sit seven orders of magnitude apart.}
  \label{fig:near-degeneracy}
\end{figure}

\section{The same drift under three metrics (C3)}

A drifting instrument does not report one number; it reports as many numbers as there are ways to score agreement. This section quantifies how far apart those readings sit \emph{on the same 100 replays}, because the gap between them is exactly the room available for motivated metric choice.

\subsection{One batch of drift, three readings}

From the F2 replay window (\S{}5.3), scoring the identical 22 drifted permutations three ways \evtag{i0r-round@3be62ed8}:

\begin{center}
\small
\begin{tabularx}{\textwidth}{>{\raggedright\arraybackslash}X>{\raggedright\arraybackslash}Xr>{\raggedright\arraybackslash}X}
\toprule
Metric & Value & Frozen gate & Reading if taken as the verdict \\
\midrule
Full-ranking agreement (exact permutation match) & 78/100 = 0.780000 & 0.99 & fail, decisively \\
Top-1 agreement & 95/100 = 0.950000 & --- & "nearly stable" \\
Pairwise order agreement & 4436/4500 = 0.985778 & --- & "almost perfectly stable" \\
\bottomrule
\end{tabularx}
\end{center}

\subsection{Why this is not a matter of metric taste}

The three numbers are not three experiments; they are one experiment under three aggregations, and their spread is structural. Full-ranking agreement decays combinatorially in the candidate count: a single adjacent transposition anywhere in a 10-alias ranking voids the entire record. The \(q^{45}\) scaling (45 candidate pairs at per-pair stability \(q\), against \(q\) for pairwise agreement) is an \emph{illustrative independence bound, not an estimate}: pair disturbances in a transitive ranking are neither independent nor identically distributed, so the exponent overstates the decay rate; the qualitative point, that record-level agreement is the fastest-decaying aggregation of the three, does not depend on independence. The observed spectrum (0.780 / 0.950 / 0.986, with a Kendall-distance spectrum concentrated at distance 1; \S{}5.3) is what near-degenerate subsets under a small noise floor produce: overwhelming pairwise stability, frequent record-level violations.

Two consequences follow. First, the frozen full-ranking gate was the strictest possible reading of the drift: the campaign's No-Go stands on the metric \emph{least} favourable to survival, which is the direction preregistration is supposed to bind. Second, and symmetrically: had the metric been chosen \emph{after} the data, two of the three readings would have supported a "nearly stable" narrative on the very replays that falsified stability at the preregistered granularity. Our protocols prohibited that substitution in advance, and this paper cites top-1 and pairwise values only as sensitivity characterisation, never as a passing verdict \evtag{docs-snapshot@a021936f}.

\subsection{Rules this yields (input to \S{}9)}

\begin{enumerate}
\item \textbf{Freeze the metric at the granularity of the downstream claim.} If the conclusion consumes only the top candidate, do not gate on the full permutation; if it consumes pairwise preferences, gate on pairwise agreement with an equivalence band. A gate stricter than the claim invites unnecessary No-Gos; a gate looser than the claim invites unsupported conclusions. Both mismatches were live risks here.
\item \textbf{Discrete exact-match gates are structurally fragile on shared endpoints.} Any readout that maps a continuous perturbation through an argmax or a permutation will convert a platform's noise floor (\S{}5.3, \S{}10) into gate-visible events at a rate the gate designer never chose.
\item \textbf{A historical footnote that generalises:} the original I-5 definition contained a continuous branch (mean absolute probability distance $\leq$ 0.02); the redesigned campaign's structured-judge readout left only the exact-equality branch applicable \evtag{ref:project-review \S3.3}. Instrument redesigns silently narrow which branches of a frozen gate remain usable; re-check the gate against the new readout \emph{before} freezing, not after.
\end{enumerate}

\section{More sampling cannot repair it (C4)}

The most natural response to an unstable readout is to average it. The original design anticipated this: its D2 protocol replaces the single-shot probability readout with \(M\) independent binary samples per measurement point, estimated with Jeffreys smoothing \(\hat r=(m+\tfrac12)/(M+1)\). This section shows why that route was closed: not by cost, but by arithmetic.

\subsection{Setup}

The frozen failure audit ran a deterministic simulation (500 replicates, seed 20260730) of the legacy D2 estimator over the I0-S repeat-task groups, using the frozen D1-S readouts as probability proxies (declared as proxies, not ground truth) and applying the frozen ranking gate (within-task Spearman median $\geq$ 0.90) to each simulated repeat \evtag{docs-snapshot@57f3389b}. Sampling noise falls as \(M^{-1/2}\); if instability were a noise problem, some \(M\) would pass the gate.

\subsection{Result}

\begin{center}
\small
\begin{tabularx}{\textwidth}{rrrrr>{\raggedright\arraybackslash}X}
\toprule
\(M\) & Observer calls & SE at \(p=0.5\) & Gate median & Gate q95 & Pass rate \\
\midrule
8 & 11,968 & 0.157 & 0.333 & 0.642 & \textbf{0/500} \\
16 & 23,936 & 0.118 & 0.304 & 0.577 & \textbf{0/500} \\
32 & 47,872 & 0.086 & 0.280 & 0.577 & \textbf{0/500} \\
64 & 95,744 & 0.062 & 0.333 & 0.632 & \textbf{0/500} \\
100 & 149,600 & 0.050 & 0.400 & 0.632 & \textbf{0/500} \\
200 & 299,200 & 0.035 & 0.447 & 0.685 & \textbf{0/500} \\
500 & 748,000 & 0.022 & 0.566 & 0.800 & \textbf{0/500} \\
\bottomrule
\end{tabularx}
\end{center}

\subsection{Why the pass rate is pinned at zero}

Between \(M=8\) and \(M=500\) the sampling standard error improves sevenfold, the achieved Spearman median climbs from 0.33 to 0.57, and the gate pass rate does not move off zero, with even the 95th percentile of the gate statistic (0.80) still below the frozen 0.90 at three-quarters of a million observer calls. (The calls column is \(M \times 1{,}496\) executed pairs; a pass rate of 0/500 bounds the per-grid-point pass probability below 0.006 at 95\% confidence, under the proxy model.) The reason is \S{}5.2's decomposition: averaging shrinks the \emph{denominator} of a resolution problem (noise), but the \emph{numerator} (the measured score separation, median \(3\times10^{-10}\) in readout units, with exact ties in a third of the groups) is a property of the sample the instrument sees. No sample size resolves a separation of zero: for the exactly tied groups, no estimator of any variance orders the pair better than a coin flip, which closes the route in principle for that third of the sample. The cost column makes the practical point redundant: the first row alone exceeds the entire I0-S campaign's call count by a factor of 3.6, and the design was rejected before cost even entered the argument.

\subsection{Boundary of the claim}

This is an exploratory measurement-error and request-scale diagnostic, not an executed experiment: it neither authorises the D2 protocol nor predicts the behaviour of \emph{redesigned} measurands (e.g. \S{}9.6's controlled-error candidates, whose gaps are constructed to be resolvable; A-S3 executed exactly that redesign, \S{}10.3). The claim is bounded by the tested grid: \textbf{for the sample this study had, no tested sampling scale (seven grid points up to \(M=500\), 500 simulated replicates each, frozen readouts as proxies) rescued the frozen ranking gate; and for the exactly tied third of the groups, no finite sampling scale can, because their separation is zero.} Extrapolation beyond the grid is an inference from the mechanism, not a measured result. A failed stability gate still cannot in general be answered with "run it bigger": the gap distribution has to be measured first, which is the point of the pilot-then-freeze rule in \S{}9.2.

\section{What the platform's own metadata can and cannot certify (C5)}

Sections 4--7 established that the instrument drifts. This section asks the audit-layer question of \emph{whether a consumer of these APIs could have known}. The answer defines the boundary of every reproducibility claim made over a shared endpoint, and it is quantifiable, because our audit chain recorded the platforms' exposed metadata verbatim (the field set of Appendix C; claims in this section are about that set on the tested providers).

\subsection{The observed facts}

\textbf{Constancy of everything visible, drift of the output.} In the F2 replay, all provider-visible metadata was constant across both windows (response \texttt{model} string 100/100, \texttt{system\_fingerprint} 100/100, both windows \texttt{null}, finish reason 100/100) while 22 of 100 byte-identical requests returned a different ranking (\S{}4.1, \S{}5.3). A metadata-constancy check of the kind audit tooling can automate would have certified this replay as clean.

\textbf{A fingerprint being present does not repair this.} The cross-provider battery (\S{}10.2) measured the fingerprint field itself, and it fails in three distinct ways \evtag{index@cf4ea148 appendix B.3}:

\begin{center}
\small
\begin{tabularx}{\textwidth}{>{\raggedright\arraybackslash}X>{\raggedright\arraybackslash}X>{\raggedright\arraybackslash}X}
\toprule
Failure form & Provider (this battery) & Measurement \\
\midrule
\textbf{Absent}: field never populated & \texttt{mistral}, \texttt{qwen} (also \texttt{qwen} in A-S1: null in 1,000/1,000) & no grouping computable \\
\textbf{Present but churning}: many values inside one window & \texttt{openai}: 19 distinct values across one day's battery; 8 distinct values within a single 37-call window of the aborted r2 run & same-fingerprint pairs agree \emph{no better} than different-fingerprint pairs (0.854 vs 0.884): the field does not predict stability \\
\textbf{Present, stable, and still insufficient} & \texttt{deepseek}: one constant, information-rich value (build, quantisation, date) & full-agreement rate under a constant fingerprint: 0.879, far from the 0.99 a replay gate requires \\
\bottomrule
\end{tabularx}
\end{center}

The strongest of the three is the last: \textbf{even a stable, semantically meaningful deployment identifier does not certify replay stability}, because the nondeterminism lives below the deployment level (batch-shape-dependent kernels, \S{}2.2). Recording the fingerprint is necessary for audit and insufficient for reproducibility; both halves of that sentence are now measured rather than argued.

\subsection{The audit proposition}

\begin{quote}
On a shared endpoint, \textbf{a model name is not a snapshot identity}, and none of the metadata fields exposed by the tested providers upgrades it into one. Any reproducibility claim whose provenance is "same model name" (or even "same model name and same fingerprint") has, as its actual scope, \emph{the visible API behaviour of that endpoint during that observation window}. Nothing further.
\end{quote}

\subsection{A three-level auditability ladder}

We propose reporting the \emph{snapshot-identity level} of any LLM evaluation as mandatory metadata, alongside temperature and seed:

\begin{center}
\small
\begin{tabularx}{\textwidth}{l>{\raggedright\arraybackslash}X>{\raggedright\arraybackslash}X>{\raggedright\arraybackslash}X}
\toprule
Level & Snapshot identity & What can be claimed & Grounding \\
\midrule
\textbf{L0} & model name only & single-window visible behaviour & this paper's campaigns (both were L0) \\
\textbf{L1} & provider-pinned snapshot/deployment ID & weights fixed; observable behaviour still not byte-stable & vendor documentation itself: model weights are fixed for a given ID while serving infrastructure changes around them, producing "minor differences in observable behavior" \citep{anthropic-model-ids}; empirically, the \texttt{deepseek} row above is an L1-like case failing a replay gate \\
\textbf{L2} & self-hosted weights hash + batch-invariant deterministic kernels + pinned hardware class and framework version & replay stability far above the shared-endpoint floor \emph{while the server is quiet}; exactness must be \textbf{verified, not assumed} & \textbf{directly measured on one stack only} (the cross-track arm below: one model, one load pattern, driver version unrecorded); the general row is a recommendation grounded in \citep{tml2025nondeterminism,vllm-batch-invariance} \\
\bottomrule
\end{tabularx}
\end{center}

The ladder's leverage lies in its middle level: L1 is where most careful practitioners believe they are safe, and it is not sufficient for replay-exact claims, by the platform's own documentation and by our measurement.

\textbf{L2, measured.} The ladder's top level was tested with the same 100-probe battery on a self-hosted arm: an open-weights judge (weights SHA-256 recorded) on a batch-invariant serving stack (vLLM 0.11.0, batch-invariant mode enabled, pinned image digest, RTX 4090-class GPU; one identity field, the driver version, was unavailable and is disclosed as missing). The arm ran on the white-box track's infrastructure and is cited by frozen artifact hash only, never pooled with any shared-endpoint number (registration in the citation ledger \evtag{repo:docs/citations.md}). \emph{Quiet} (serial, five passes): full-ranking agreement 0.9872 over pass pairs with well-formed replies (0.928 under the frozen criterion, which counts the judge's occasional malformed rankings as disagreement), far above the 0.740--0.879 shared-endpoint range. \emph{Under batch pressure} (eight concurrent background streams): 0.8925 clean (0.830 frozen), back inside that range; per-pair disagreement rose from 1.28\% to 10.75\%, a factor of 8.4. Two deliberately narrow readings follow. First, the mechanism account's prediction of exact replay (\S{}9.1) did not hold as stated: batch-invariant kernels were necessary but not sufficient on this stack, so L2's claim is earned by verification, not by flag. Second, batching pressure alone \emph{suffices} to reproduce shared-endpoint-magnitude instability on one model and one stack; whether it is \emph{the} cause of the shared-endpoint floor remains unidentifiable from these data, and this paper does not claim it.

\subsection{Four further boundary forms, found by trying to instrument the audit}

Preparing the cross-provider battery required verifying, per provider, that each audit-relevant request field actually functions. That verification (three capability probes and one throughput probe, 270 calls in total) surfaced audit boundaries of its own kind \evtag{index@bee9bc61 appendix B.2--B.3}\evtag{repo:docs/reports/a\_s2\_capability.md, a\_s2\_capability\_r2.md, a\_s2\_capability\_r3.md, a\_s2\_throughput.md}:

\begin{enumerate}
\item \textbf{Documentation is not an audit source.} Documented parameter handling contradicted measured behaviour four times: fields documented as silently ignored were \textbf{rejected} (HTTP 400/422 on \texttt{seed} and on a \texttt{response\_format} type the vendor's compatibility guide lists as ignored); a documented hard cap was \textbf{silently ignored} (a completion-token limit exceeded 39-fold by a reasoning-mode default); a field documented as populated returned \texttt{null}. The direction varied; the presence did not. Probe per provider, per model, per field.
\item \textbf{A platform can forbid determinism at the protocol level.} One provider rejects \texttt{temperature=0} outright (only 0.6 accepted): an audit boundary no client-side discipline crosses.
\item \textbf{Rate-limit observability is bimodal.} One provider publishes its quota precisely enough to predict the first throttled call; two return no rate-limit headers at all, and observed 429s carried no \texttt{retry-after}. Load, and therefore batch shape (the very variable the nondeterminism mechanism runs on), is invisible on half the platforms.
\item \textbf{Aliases move under audit.} Dated snapshot names resolve stably in the \texttt{model} response field; undated aliases are retireable pointers (one provider retired its rolling aliases mid-2026; another's binding is explicitly unversioned). The audit trail records the motion only if the response-side model string is logged per call, a mandatory audit field (\S{}9.7).
\end{enumerate}

\subsection{Implication for reporting}

The reporting consequences of this section are checklist items 1--4 (Appendix D): snapshot-identity level, response-side model distribution, fingerprint status measured over the run, and probed parameter handling. Absent these, a diverging replication cannot even be classified as instrument drift versus platform drift; with them, the divergence becomes attributable, which is all an audit can promise on L0/L1 infrastructure.

\section{Design rules for preregistered LLM evaluation (C6)}

Nothing here is advice in the abstract: each rule names the failure it prevents, every failure named was purchased and audited in \S{}\S{}4--8, and Appendix A.11 maps each rule to its evidence.

\subsection{Lock the snapshot before trusting any stability gate ($\leftarrow$ C5, M3)}

An I-5-class gate on an L0 endpoint tests a proposition no one controls. Prefer self-hosted open weights, then provider-pinned snapshots (L1, which fix the weights but not the observable behaviour; \S{}8.1), then shared endpoints only with gates over aggregated readouts (\S{}9.5). An earlier version called L2 replay-exactness a property "by construction"; the measured arm corrected it (\S{}8.3). \textbf{Lock the snapshot, then measure the served path's stability under representative load, rather than inferring it from the serving stack's guarantees.}

\subsection{No gate freezes on a never-measured quantity ($\leftarrow$ C1, C4)}

Both campaign-ending gates were frozen on quantities no one had piloted, and neither threshold was reachable by any instrument of the audited class (\S{}4.1, \S{}10.1): they tested nothing, at full ceremony and real cost. The repair is a fixed pipeline: \textbf{pilot} the noise floor and gap distribution, \textbf{simulate} the gate's operating characteristic over healthy- and degraded-instrument models (healthy $\geq$ 0.9, degraded $\leq$ 0.1), \textbf{freeze} only if discrimination is adequate. This project possessed exactly the simulator required (\S{}7) and used it post mortem; the pilot that finally exposed I-5R's unreachability cost CNY 3 against 52,988 audited attempts.

\subsection{Condition ranking gates on informative pairs ($\leftarrow$ M2)}

A ranking gate's denominator must contain only pairs the instrument could in principle resolve: pilot-measured gap above a resolution bound \(\delta\) set from the noise floor (0.0067 median, 0.080 p95; \S{}5.2), ties and near-degenerate pairs excluded by preregistered rule before any data is seen. "Mostly below detection limit" is reported as a property of the sample (84\% of groups here), not instrument failure. Where possible, check the bound from outside the loop with a constructed-error battery stratified by \emph{both} error magnitude and error mechanism: 340 calls showed separation governed by error type, not magnitude (\S{}10.3), so a single pooled detection-limit curve can be composition-confounded and is not sufficient on its own.

\subsection{Prefer continuous readouts with equivalence bands ($\leftarrow$ C3)}

Exact-match gates over discrete structures amplify the noise floor combinatorially (pairwise 0.986 becomes record-level 0.780; \S{}6.2). Gate on continuous aggregates against a preregistered equivalence band, or decompose rankings into explicitly aggregated per-candidate or pairwise scores, so one token flip moves one comparison, not the whole record. The frozen metric must sit at the granularity of the claim it licenses (\S{}6.3).

This is not a licence to swap a strict gate for a friendlier statistic. When the downstream use genuinely consumes a full ranking, the operational prescription is: keep the full-ranking deliverable but make the \emph{aggregate}, not the single call, the measured object. Concretely, (i) freeze \(k\), the aggregator (Borda or median rank with a deterministic tie rule), and the gate before execution; (ii) set the equivalence band from the piloted floor (for example, aggregated disagreement at or below the pilot's 95th percentile); (iii) decide by the TOST convention, the whole interval inside the band (\S{}2). Sizing on this study's own measured spectrum says the price is modest: simulating Borda-of-\(k\) readouts under the A-S1 same-day Kendall spectrum (calibrated so \(k=1\) reproduces the measured pair agreement of 0.7985), two independent aggregated batteries agree exactly at rate 0.94 for \(k=3\), 0.98 for \(k=5\), and 0.999 for \(k=9\), with within-one-transposition agreement at 0.999 by \(k=5\) \evtag{repo:docs/reports/aggregation\_prescription.json}. A full-ranking gate that was unreachable per call (\S{}5.3) becomes passable at five to nine calls per measurement point, which is the difference between abandoning the granularity and retaining it at a known, preregistered cost. The simulation shares \S{}7's evidentiary tier: it sizes a design and authorises nothing (\S{}11.4).

\subsection{Put aggregation inside the measurement definition ($\leftarrow$ M3, \S{}10.1)}

On L0/L1 infrastructure a single call is one draw from the platform's nondeterminism distribution. Fix \(k\), the aggregator, and the gate before execution, and keep every draw; the forbidden variants (replay-until-agreement, post-hoc majority votes, dropping discordant draws) condition on the verdict. A-S1 bounds what aggregation buys: the floor is instantaneous (0.805 versus 0.800, replicated), and scheduling remedies offer nothing that can be planned in advance: where a day did differ (r2's day 2), nothing observable marked it beforehand (\S{}10.1).

\subsection{Make the measurand exist ($\leftarrow$ M2, upstream)}

No instrument rule repairs quantities degenerate by construction. Manufacture candidate sets to be resolvable: controlled error injection, difficulty near accuracy 0.5, mixed-correctness coverage above a preregistered floor. The values this study ran with (0.781, 0.163; \S{}5.2) should have been disqualifying at design time; the \S{}9.2 pilot would have caught them.

\subsection{Report against a fixed checklist (Appendix D)}

The recurring failure behind M1--M3 is unstated instrument status. Appendix D gives the checklist this study's evidence supports, phrased so each item is reportable from data an audit chain already collects (\S{}3.5): adopting it costs logging, not methodology.

\subsection{Fail closed: a guard that cannot parse must refuse ($\leftarrow$ \S{}12.1)}

Both of this study's aborted windows trace to guards that defaulted to allow on unrecognised input (a wire-shape parameter, a scheduling check; Appendix A.10). Unparseable means refused, and any parameter that load-bears for the science is verified at the wire or response level (\S{}8.4), never assumed from SDK acceptance. We disclose both incidents because a methods paper whose tooling failed open twice, caught it, and re-froze rather than patching silently is the checklist argument in miniature.

\section{Prospective supplements: decomposing, generalising, and calibrating the floor}

The legacy campaigns left three attributions open: whether the replay disagreement of \S{}5.3 was a \emph{cross-day} effect or an instantaneous one; whether it was a property of one endpoint or of the platform class; and whether the readout can resolve true differences of known size at all, a question the legacy data could pose only in the instrument's own units (\S{}5.2). All three were answered with new, preregistered, separately archived supplements, never pooled with the historical data, which enters below only as a qualitative reference line. The main-text claims were frozen before any supplement ran, and none moved a threshold: the supplements sharpen attribution and scope, the role prospective replication should play in a measurement paper.

\subsection{A-S1: same-day/cross-day variance decomposition}

\textbf{Design} (preregistration frozen at commit \texttt{ffbeb95} before execution; descriptive, no pass/fail gate). The 100 frozen judge request bodies of the F2 window (\S{}5.3) were replayed as a fixed probe battery: five windows per day with a $\geq$ 2 h spacing rule, on two days, \(10\times100 = 1{,}000\) calls against the same endpoint and dated model snapshot as the historical window. Interpretation bands were frozen in advance: same-day/cross-day median difference \(|d|<0.03\) reads as a nondeterminism floor, \(d\ge0.10\) as a systematic cross-day component.

\textbf{Result.} Coverage was complete (1,000/1,000 valid readouts, zero transport failures, \texttt{system\_fingerprint} null in 1,000 of 1,000) and the decomposition is unambiguous \evtag{index@cf4ea148 appendix A.3}\evtag{repo:docs/reports/a\_s1\_samedayvar\_r1.md}:

\begin{center}
\small
\begin{tabular}{ll}
\toprule
Quantity & Value \\
\midrule
median full-agreement, 20 same-day window pairs & \textbf{0.805} \\
median full-agreement, 25 cross-day window pairs & \textbf{0.800} \\
difference \(d\) & \textbf{0.005} $\rightarrow$ \textbf{nondeterminism floor} \\
top-1 agreement medians (same-day / cross-day) & 0.950 / 0.960 \\
pairwise agreement medians & 0.9859 / 0.9860 \\
\bottomrule
\end{tabular}
\end{center}

The historical 0.78 sits \emph{inside} the same-day distribution (range 0.73--0.83). Twenty-four hours contribute nothing measurable: what \S{}5.3's window observed was the endpoint's instantaneous nondeterminism, reachable twice in one afternoon. The metric spectrum reproduces at the same three magnitudes as the frozen window (0.780/0.950/0.986 historically; 0.805/0.950/0.986 here) without any pooling: two independent measurements of one floor.

\textbf{Dependence-aware uncertainty.} The 45 contrasts share ten windows, so the frozen-band reading is re-checked with the \emph{window} as sampling unit \evtag{repo:docs/reports/dependence\_reanalysis.json}. A cluster bootstrap over windows (10,000 replicates; five per day, resampled with replacement; self-pairs dropped) puts the same-day median at 95\% interval [0.77, 0.83], the cross-day median at [0.76, 0.82], and \(d\) at [-0.03, 0.04]; 87\% of replicates fall inside the band. The exact reference is sharper. Over all 252 balanced relabelings of the ten windows into two pseudo-days, \(|d|\) at least as large as observed arises in 98\% of splits, and no relabeling produces \(|d| > 0.025\): the decomposition is a property of the window collection, not of the day labels.

\textbf{Consequences for the argument.} \S{}5.3 was renamed by this result, and \S{}9.5 (prospective aggregation) gains its quantitative basis: the floor is per-call, so \(k\)-fold aggregation attacks it. Two days cannot falsify the existence of a stabler day somewhere; they do show that cross-day variation was indistinguishable from the same-day floor whenever we looked (\S{}11.4).

\textbf{Replication, and the day that stood out (r2).} A second preregistered round (frozen at commit \texttt{29b5962}) ran the same frozen battery over five days spanning 2026-08-14 to 08-26: 2,500 of 2,500 valid readouts, fingerprint null throughout \evtag{index@cf4ea148 appendix D.3}\evtag{repo:docs/reports/a\_s1\_samedayvar\_r2.md}. Same-day median 0.8150 (50 pairs) versus cross-day 0.8100 (250 pairs), \(d = 0.0050\): the frozen replication rule fired. The rounds remain separate units, never pooled; "seven days of coverage" describes scope, not a sample. With 25 windows the dependence-aware intervals tighten: \(d\) at [-0.01, 0.03], entirely inside the band (99.8\% of replicates), and no relabeling among 10,000 seeded balanced ones produces \(|d| > 0.02\) \evtag{repo:docs/reports/dependence\_reanalysis.json}. The round also asked what r1 could not: whether the days \emph{themselves} are alike. They are not. Per-day same-day medians run 0.8250 / 0.7800 / 0.8000 / 0.8200 / 0.8250, a spread of 0.0450 against the 0.03 band, and the frozen reading fired: \emph{a day stood out} (day 2, the least stable). Both findings stand because they are different quantities: the cross-day median never leaves the floor, while the per-day floor itself wanders beyond the band. "Measure on a good day" is therefore not even well-posed prospectively: nothing recorded distinguishes day 2 in advance, and its pre-flight canary passed all four assertions. Every waiting-related statement in this paper stays bounded to the days sampled.

\textbf{Disclosure.} The first A-S1 attempt (v1) is retained as a failed instrument window (100 calls, 100 responses, zero valid readouts: an SDK-shape parameter never entered the wire body; \S{}9.8, \S{}12.1), permanently excluded and archived separately. In r2, the first day-4 attempt (2026-08-20) is likewise retained, voided, in its own archive: two executor processes concurrently wrote one run directory, window 2 received 285 requests with 21 probes carrying conflicting readings, and day 4 re-ran four days later under a new run group and a runbook guard against concurrent executors (\S{}12.1). One governance deviation ships with it: the round's authorisations were pre-signed for seven candidate dates (the researcher was travelling), trading the daily human checkpoint for the frozen runbook's machine preflights; one date expired unused. One limitation is inherited by construction: byte-equivalence of the replayed bodies with the F2 originals is asserted by construction, not verified against archived wire bytes, because the legacy archive records SDK-parameter shape (\S{}11.5).

\subsection{A-S2: the floor is a platform-class property}

\textbf{Design} (preregistration frozen at \texttt{cdc012f}; descriptive; per-provider analyses never merged). Four providers of mutually distinct families received the same 100-probe battery (\texttt{deepseek-v4-flash}, \texttt{mistral-small-2603}, \texttt{gpt-4.1} returning dated snapshot \texttt{gpt-4.1-2025-04-14} in every response, and \texttt{qwen3.7-plus-2026-05-26} as the A-S1 anchor) through byte-equivalent requests differing only in \texttt{model} and base URL: five windows $\times$ two days $\times$ four providers = 4,000 calls, same-day spacing 2.02--2.04 h, cross-day gap 24.14 h. Three frozen readings: a platform-class test (max per-provider same-day median < 0.95), a per-provider decomposition (bands as in A-S1), and a fingerprint-predictiveness test.

\textbf{Result.} Coverage 4,000/4,000 responses, zero transport failures \evtag{index@bee9bc61 appendix B.3}\evtag{repo:docs/reports/a\_s2\_xprovider\_r3.md}:

\begin{center}
\small
\begin{tabularx}{\textwidth}{lr>{\raggedright\arraybackslash}X>{\raggedright\arraybackslash}X>{\raggedright\arraybackslash}X}
\toprule
Provider & valid rate & median same-day agreement & \(d\) (same - cross) & decomposition \\
\midrule
\texttt{deepseek} & 0.990 & \textbf{0.879} & 0.000 & nondeterminism floor \\
\texttt{mistral} & 1.000 & \textbf{0.740} & 0.000 & nondeterminism floor \\
\texttt{openai} & 1.000 & \textbf{0.850} & 0.000 & nondeterminism floor \\
\texttt{qwen} & 1.000 & \textbf{0.775} & -0.005 & nondeterminism floor \\
\bottomrule
\end{tabularx}
\end{center}

All four providers entered the platform-class test (valid rate $\geq$ 0.95 precondition), and its frozen reading fired: \(\max = 0.879 < 0.95\) $\rightarrow$ \textbf{platform-class property}, a verdict name whose scope is exactly what was measured: these four providers, this 100-probe structured-ranking battery, these observation windows --- valuable evidence about shared serving infrastructure, not a census of it. The window-level bootstrap and exact relabeling of \S{}10.1, applied per provider, give \(d\) intervals within $\pm$0.03 for every provider (in-band fractions 0.91--0.95; all 252-split relabeling probabilities 1.0; no relabeling exceeds \(|d| = 0.03\)) \evtag{repo:docs/reports/dependence\_reanalysis.json}. No provider approaches replay-exactness; every provider's cross-day behaviour equals its same-day behaviour; the ordering readout's stability differs by provider (0.74--0.88) but its character does not. The fingerprint field completes the picture in three failure forms, absent / churning / stable-but-insufficient, measured in \S{}8.1.

\textbf{Reading discipline.} These are stability comparisons, not quality comparisons: the four models sit at different capability tiers by design choice, and no cross-provider number above compares judging accuracy (\S{}11.4). The \texttt{qwen} column is an independent re-measurement of A-S1's endpoint in new windows: 0.775 against A-S1's 0.805, read side by side and never pooled. One verbatim-reproducible failure is disclosed rather than patched: a single probe on \texttt{deepseek} returned a bare JSON array instead of the required object in all ten windows, byte-identical every time, scored as invalid per the frozen rule, ten out of ten, on both days.

\textbf{Excluded windows.} Two aborted Day-1 attempts (238 and 618 calls) are retained for disclosure and enter no denominator: the first exposed a rate-limit quota that a frozen premise had assumed away, the second a fail-open scheduling guard (\S{}9.8); both triggered full re-freezing rather than in-place amendment (\S{}12.1).

\subsection{A-S3: a controlled-error battery puts resolution outside the loop}

Everything above measures the observer against its own readouts: \S{}5.2's gap distribution and noise floor come from the same instrument, so "the gaps were too small to resolve" is, on the legacy data alone, an operational diagnosis rather than an independently identified fact. A-S3 constructs the measurand instead.

\textbf{Design} (preregistration frozen at \texttt{c145f0c}; single provider, single session; not a stability measurement). The 25 task stems of the I0-R battery were reused (re-derived in exact rational arithmetic, 25 of 25), but every candidate was rebuilt: at each of 85 (task $\times$ visible-prefix) variants, one correct candidate plus nine \emph{slip} candidates from a five-type catalogue of human-plausible errors, injected at rotated step positions, so the relative error \(\varepsilon\) of every slip is known by construction before any call. All ten candidates share identical wording; arithmetic is the only signal, and the plan builder refuses to execute on a battery-hash mismatch. Four passes over the 85 variants gave 340 calls, each returning one full ranking; a \emph{judgment} is whether the correct candidate strictly outranks a given slip candidate (85 $\times$ 9 $\times$ 4 = 3,060 judgments; chance 0.5) \evtag{index@cf4ea148 appendix C.3}\evtag{repo:docs/reports/a\_s3\_detection.md}.

\textbf{Result.} Coverage 340/340 valid readouts, zero transport failures, \texttt{system\_fingerprint} null in 340 of 340 (L0):

\begin{center}
\small
\begin{tabularx}{\textwidth}{>{\raggedright\arraybackslash}Xrrl}
\toprule
\(\varepsilon\) bin & judgments & separation & Wilson 95\% \\
\midrule
below \(10^{-3}\) (outside the main verdict) & 96 & 0.7604 & [0.6661, 0.8347] \\
\([10^{-3}, 10^{-2})\) & 556 & 0.7914 & [0.7556, 0.8231] \\
\([10^{-2}, 10^{-1})\) & 932 & 0.8069 & [0.7803, 0.8309] \\
\([10^{-1}, 10^{0})\) & 1,068 & 0.9860 & [0.9770, 0.9915] \\
\([10^{0}, \infty)\) & 408 & 0.9020 & [0.8692, 0.9272] \\
\bottomrule
\end{tabularx}
\end{center}

Every group is \emph{resolved} (Wilson lower bound above 0.5), so the frozen main verdict is the fourth of the preregistration's four named outcomes: \textbf{the battery did not bracket the detection limit}; it certifies only that the limit lies below \(10^{-3}\) relative error for constructed slips.

\textbf{The within-type contrast governs.} Error size and slip type are correlated by construction, so the preregistration ordered a same-type contrast and declared it decisive over the pooled table. It reverses the naive reading: in four of five slip types, separation does \emph{not} rise with \(\varepsilon\), and the largest-error cell of one type (off-by-one at \(\varepsilon \ge 1\), n = 36) sits at 0.333, below chance. The pooled rise across bins is mostly \emph{composition} (which slip types populate which bin), not error magnitude. What the battery certifies is a discriminator sensitive to error type and surface form, not a meter that orders candidates by the size of their error. Two further disclosures: replay disagreement across the four passes was 19 of 765 pairs (0.0248), minutes apart and therefore not comparable to any same-day agreement number in \S{}10.1--\S{}10.2; and the battery's scope is fixed by its construction, with uniform rewritten wording (a deliberate difference from the I0-R candidates, whose styles vary), a prefix-length parameter that does not inherit the archived progress scale, one provider, one task family. Extrapolating its detection limit to the original I0-R candidates would need a separate argument (\S{}11.1).

\begin{figure}[t]
  \centering
  \includegraphics[width=\linewidth]{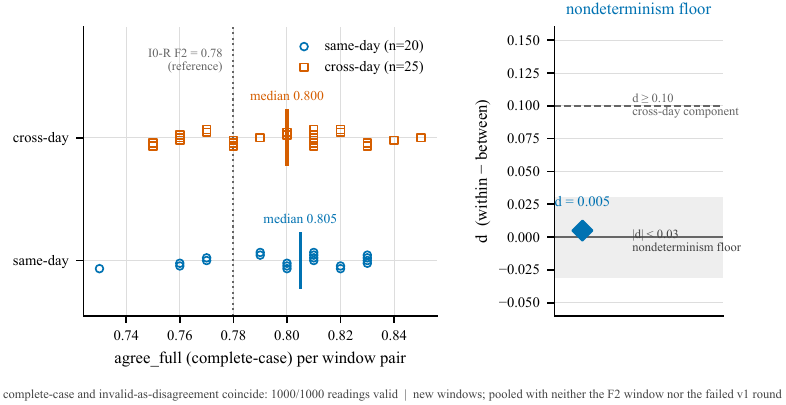}
  \caption{A-S1, the same-day/cross-day decomposition of full-ranking
  agreement (\S{}10.1): 20 same-day and 25 cross-day window-pair
  agreements from ten 100-probe windows, with medians and frozen bands;
  the historical F2 value (dotted) sits inside the same-day distribution.
  New windows, pooled with nothing.}
  \label{fig:as1}
\end{figure}

\begin{figure}[t]
  \centering
  \includegraphics[width=\linewidth]{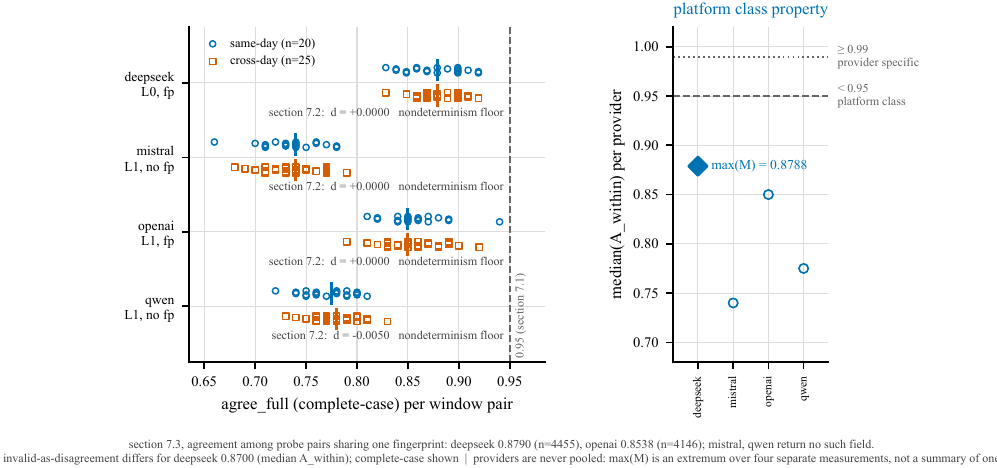}
  \caption{A-S2, cross-provider stability of the same battery (\S{}10.2):
  per-provider window-pair agreement distributions and medians, never
  merged; none approaches the 0.99 a replay gate would need. Stability
  comparison only; capability tiers differ by design, and nothing here
  compares judging accuracy.}
  \label{fig:as2}
\end{figure}

\section{Threats to validity and the boundary of claims}

\subsection{External validity}

The mechanism evidence (M1, M2) comes from one observer family (\texttt{qwen3.7-plus}) on one shared endpoint, over one task family (synthetic, deterministic rational arithmetic with exact canonical answers), one prompt template, and one language. A-S2 extends the \emph{stability} characterisation to four providers across three jurisdictions, and A-S1 gives the same-day noise model for the primary endpoint; neither extends the M1/M2 \emph{mechanism} measurements (mapping bias, degeneracy decomposition) beyond their original instrument, and we claim nothing broader for them. The A-S3 detection-limit numbers are similarly bounded: they hold for constructed slip candidates with uniform rewritten wording on one provider in one session, and do not transfer to the stylistically heterogeneous I0-R candidates without a separate argument; its four-pass replay disagreement (0.0248, minutes apart) is not a same-day stability measurement and is never compared with A-S1/A-S2 numbers. The L2 control point has now run (\S{}8.3): one open-weights model on one batch-invariant serving stack, under two load conditions, executed on the white-box track's infrastructure and cited by frozen artifact hash. Its boundaries are inherited by every statement that uses it: a single model, stack, GPU class, and concurrency setting; one identity field (driver version) unrecorded on that stack, disclosed; and its design supports sufficiency (batching pressure can produce shared-endpoint-magnitude instability) but not attribution (that the shared-endpoint floor \emph{is} batching), which this paper accordingly never claims.

\subsection{The measurand was ours}

The near-degeneracy at the heart of M2 was manufactured by our task design (candidate accuracy 0.781, mixed coverage 0.163; \S{}5.2), not discovered as a law of LLM-as-judge practice. Deployed judge pipelines with well-separated candidates need not exhibit ranking collapse of this severity. The claim we defend is invariant to that: \textbf{a ranking gate is only meaningful relative to the measured gap distribution of its measurand}, and a study that has not measured its gap distribution does not know which regime it is in. We were in the degenerate regime and did not know until the failure audit; that ignorance, not the degeneracy itself, is the transferable hazard.

\subsection{What this paper does not claim}

None of the results here bear on the internal states, representations, or semantic dynamics of the observed models; every quantity is an operationalisation of external measurement over visible records (\S{}3.1). This paper is not a test of any theory of model internals; the project's white-box track addresses those questions separately, under its own preregistration, and no data flows between the two. Nor is any result a service-quality assessment of a named provider: the behaviours documented are consistent with the public engineering properties of shared inference services (dynamic batching without batch-invariant kernels), our provider exclusions were protocol- or account-level (\S{}8.4), and the capability tiers of the A-S2 models were deliberately not equalised (\S{}10.2). Finally, this paper does not conclude that LLM-as-judge evaluation is unusable; it concludes that judge \emph{instruments} have measurable noise floors, gap distributions, and audit boundaries, and that gates written without measuring them test nothing.

\subsection{Statistical limitations}

The I0-S ranking verdict rests on 31 valid groups of 32 (one group's repeat ranking is constant, its Spearman undefined, reported as undefined and not imputed to 0 or 1). The D2 simulation (\S{}7) uses frozen D1-S readouts as probability proxies, declared as proxies; its conclusion is about the frozen gate on this measurand, not about redesigned measurands. The legacy 136-pair diagnostic behind M1's \(b_{\mathrm{map}}\) quantiles is exploratory (preregistration-ineligible) and is used only for mechanism characterisation, never as gate evidence. The \S{}9.4 aggregation sizing simulates on the measured A-S1 same-day spectrum of one endpoint and one battery, under an adjacent-transposition surrogate calibrated at \(k=1\); its \(k\) values size a design on that floor and transfer to other endpoints only after their own pilots. The supplements' medians summarise window-pair sets that share windows and are not independent, so \S{}10 reports window-level cluster bootstrap intervals and balanced day-relabeling references alongside the frozen bands \evtag{repo:docs/reports/dependence\_reanalysis.json}. The same-day/cross-day arms sample two days (r1) and five days (r2), never pooled; cross-day statements are bounded by those designs and never extrapolated to "any future day", a boundary r2 made concrete by measuring a day whose own floor stood out (\S{}10.1). The A-S2 judge shares a model family with the generator of the historical candidates (\texttt{gpt-4.1} judging \texttt{gpt-4o-mini} outputs), a self-preference confound we disclose; it can shift \emph{which} permutation a judge prefers, not \emph{whether the same request reproduces the same permutation}, which is the only quantity A-S2 reads.

\subsection{Preregistration deviations, disclosed}

Two, both bounded. In the second legacy campaign, the calibration-stage freeze postdates two development-stage results; the record marks it a preregistration-timing deviation, and this paper accordingly never describes I0-R as fully pre-frozen (the F2 replay protocol itself, and its 0.99 gate, were frozen before the replay window opened). In A-S1, byte-equivalence between the probe battery and the historical F2 wire bodies is asserted by construction rather than verified against archived bytes, because the legacy archive records requests in SDK-parameter shape (\S{}10.1); every stability statistic in \S{}10 is internal to the new windows and unaffected. The supplements' own aborted windows and their causes are disclosed in \S{}12.1 and were handled by re-freezing, not amendment.

\section{Disclosure, data, and reproducibility}

Every number in this paper resolves to a file inside a frozen, hash-manifested archive; every archive is listed below with its package SHA-256; and every run that was excluded from analysis is disclosed here with the reason and the disposition. The index that governs all of it is itself version-pinned \evtag{index@cf4ea148}.

\textbf{Evidence packages.} Legacy campaigns (sealed 2026-08-03):

\begin{center}
\small
\begin{tabularx}{\textwidth}{>{\raggedright\arraybackslash}X>{\raggedright\arraybackslash}Xrr>{\raggedright\arraybackslash}X}
\toprule
Package & SHA-256 & Members & Bytes & Role in this paper \\
\midrule
\texttt{phase1-oldD1.zip} & \texttt{\seqsplit{859ba16e971c3ca2f87e7dbc75106d00340852801acc7cdf1a6d97714c68ae56}} & 83 & 2,989,738 & failure mode M1; legacy ledgers; excluded windows \\
\texttt{phase1-i0s-fresh.zip} & \texttt{\seqsplit{e41bb822aebb310e38f2618df20b0b942d1603420365785e69100f01bc9c8e44}} & 41 & 2,215,094 & failure mode M2; I-3S primary evidence; D2 simulation \\
\texttt{i0r-round.zip} & \texttt{\seqsplit{d0cff5cdcd4d3005338f34d9a077f8a2e552311a2f636762c922460f4b65040d}} & 458 & 66,243,938 & failure mode M3; I-5R primary evidence; metric spectrum \\
\texttt{docs-snapshot.zip} & \texttt{\seqsplit{87179236958ec882f091a57db1ceafcfd07b44d202889e3f485c67549114e24c}} & 204 & 1,112,797 & preregistration texts, gate definitions, closeout reports \\
\bottomrule
\end{tabularx}
\end{center}

Prospective supplements (this study's new windows, sealed 2026-08-05/-26):

\begin{center}
\small
\begin{tabularx}{\textwidth}{>{\raggedright\arraybackslash}X>{\raggedright\arraybackslash}Xr>{\raggedright\arraybackslash}X}
\toprule
Package & SHA-256 & Members & Role \\
\midrule
\texttt{a-s1-r1.zip} & \texttt{\seqsplit{196ab2e72cd10c72324ed269163feb39dd95a5ca2c9da11e644b01e1e7670d3a}} & 39 & A-S1 primary evidence (\S{}10.1) \\
\texttt{a-s1-v1-failed.zip} & \texttt{\seqsplit{4caac1c8e270b82ab47dd92a17e513d17609a5ec2e8634332781a56f27a93def}} & 19 & A-S1 v1 failed instrument window (disclosure only) \\
\texttt{a-s2-probes.zip} & \texttt{\seqsplit{ebde3c776727e774444528ed1896c021347066c1a40d371bab8ca43ee6d81d88}} & 50 & capability \& throughput probes (instrument calibration) \\
\texttt{a-s2-abandoned.zip} & \texttt{\seqsplit{f5f4a64f3e8d0531a8daf528bd3a942869003dc0d34bbd92fa15619195e64115}} & 35 & A-S2 v1 (never run) and r2 aborted window (disclosure only) \\
\texttt{a-s2-r3.zip} & \texttt{\seqsplit{07ae32a50776a1e19b5bf7ed1ceebc334bf94d4c5981eab78c5e17037445e365}} & 30 & A-S2 primary evidence (\S{}10.2) \\
\texttt{a-s3.zip} & \texttt{\seqsplit{a333e15fe5e3b5712f40a964860f4bad155d58d498fce3de09580e82e91d647f}} & 7 & A-S3 controlled-error battery, primary evidence (\S{}10.3) \\
\texttt{a-s1-r2.zip} & \texttt{\seqsplit{682b9b73e138c1268b68f233651fcba10c43109bc3df81d64a51e574d6c5348a}} & 70 & A-S1 r2 five-day replication, primary evidence (\S{}10.1) \\
\texttt{a-s1-r2-voided.zip} & \texttt{\seqsplit{35a5450d33f74a1511dbc0c67c1040372fff0d6d149addd7f24cd774028b80ac}} & 13 & A-S1 r2 voided day-4 first attempt (disclosure only) \\
\bottomrule
\end{tabularx}
\end{center}

\subsection{Excluded data (retained for disclosure; never in any denominator)}

Nothing was deleted, edited, or repaired anywhere in either campaign or in the supplements; "excluded" always means \emph{retained, disclosed, and kept out of every denominator}, with costs left on the ledger. Eleven exclusion events span the study, from Round 0's duplicate candidates through the supplements' three aborted or voided windows; the full register (event, disposition, standing obligation) is Appendix A.10, sourced from \evtag{index@bee9bc61} \S{}4 and appendices A--D and the per-run ledger \evtag{repo:docs/costs.md}.

\subsection{Request-attempt ledger}

These figures exist so that no reader mistakes audit volume for sample size; no scope was ever pooled into another, and none of the totals counts successful or billable calls. The legacy workspace logged 4,088 attempts in the D1 group and 3,314 in the fresh I0-S group (phase-1 total 7,402), plus 45,586 in the I0-R campaign: 52,988 attempts, of which 52,574 received a provider response, 413 failed in transport, and one carries neither event. That residual category exists by construction: the cap counter reserves \emph{before} each network attempt (\S{}3.5), so an attempt interrupted between reservation and outcome is counted without an event record, and the arithmetic closes at 52,574 + 413 + 1 = 52,988 \evtag{index@bee9bc61 \S{}8}. The supplements add 9,467 attempts: A-S1 with its failed v1 window (1,104), the A-S2 probes, aborted windows, and battery (5,126), A-S3 (340), and A-S1 r2 with its voided day (2,897). The four primary batteries completed 1,000/1,000, 4,000/4,000, 340/340, and 2,500/2,500 with zero transport failures (\S{}10); per-window detail ships in the supplement's ledgers.

\subsection{Reproduction package and access tiers}

The supplement is a purpose-built public-layer package, not a copy of the archives. Two tiers, decided before writing began:

\begin{center}
\small
\begin{tabularx}{\textwidth}{l>{\raggedright\arraybackslash}X}
\toprule
Tier & Contents \\
\midrule
Public & every manifest; all derived data; reports; sealed configuration snapshots; cost ledgers; preregistration texts (frozen commits); figure-rebuild code and tests; for A-S2, provider response headers sanitised at capture time by a credential denylist and secret-shape heuristic \\
Restricted & \texttt{raw/api\_events.jsonl} (verbatim provider responses) for every run; one operator-side launcher script that embeds a local account path (frozen bytes are never edited, so it ships hash-only whole) \\
\bottomrule
\end{tabularx}
\end{center}

The public package includes the SHA-256 of every restricted-layer file, so the integrity chain is verifiable end-to-end without access to raw responses. The complete run-directory index (110 runs with archive, manifest anchor, and standing classification) ships in the supplement as the store's own searchable index; the packages table above is its compact summary. Every figure in this paper rebuilds from the public layer alone with one command and zero network calls; the package asserts its own byte size and re-verifies its own hashes at build time. Five further figure views (progress-stratified stability, the metric-sensitivity bars, the sampling-scale panels, the drift-locality panels, and the audit-ladder diagram) rebuild with the same command and ship in the supplement; each duplicates numbers or structure kept in the main text. Raw provider responses are held as sensitive research data, available on request, subject to a provider terms-of-service review at publication time.

\subsection{A provenance gap, honestly registered}

Nineteen phase-1 runs predate the repository's git history and reference configuration files \emph{outside} their run directories by absolute path; those files changed later (28 items). In-run evidence shows zero mismatches, but the exact bytes those runs executed under cannot be reproduced, so every conclusion drawn from them is capped at exploratory-diagnostic status. The two runs sealed \emph{inside} their run directory with exact hash agreement (the I0-S canary and full run) are the only phase-1 runs cited as primary evidence \evtag{index@bee9bc61 \S{}6}.

\subsection{Cost ledger}

Reported so that the scale of evidence is not mistaken for scale of spend, and because cost asymmetry is itself a finding: the entire F2 refutation window cost CNY 0.296479056 \evtag{i0r-round@e8420c2f}. The legacy campaigns closed at CNY \textbf{30.868827240} (the sealed ledger through Stage 3-E4 \evtag{docs-snapshot@c01bb1c6} plus that F2 window, the only later legacy spend); the I0-S full run alone cost 1.742772 \evtag{docs-snapshot@7b8f55c0}. The supplements added $\approx$ 37.73: A-S1 r1 $\approx$ 2.973, A-S2 probes and aborted windows $\approx$ 4.63 (sunk, disclosed), A-S2 r3 $\approx$ 20.623, A-S3 $\approx$ 0.932, and A-S1 r2 $\approx$ 7.43 plus its voided day-4 attempt $\approx$ 1.145 \evtag{repo:docs/costs.md}. Provider billing statements remain authoritative. Every live window was authorised in writing before execution, against a per-line budget envelope, with a hard request cap wired into the executor; the itemised per-window ledger and the authorisation ledger ship in the public layer.

\section{Conclusion}

Two preregistered audits of black-box LLM observers on shared endpoints terminated at their instrument gates, with execution records at ceiling and scientific verdicts far below threshold. The dissection is compact. Transport perfection implies nothing about measurement reliability (C1). The failures decompose into a biased readout, near-degenerate score separations, and an instability that byte-identical inputs could not cross, amplified by a permutation readout into one observable (C2). Metric substitution did not repair them (C3), nor did scale on the tested sampling grid (C4). The platform's own metadata cannot certify what reproducibility claims need, even when the fingerprint field is populated and stable (C5). Each mechanism converts into an enforceable design rule, summarised as a reporting checklist whose every item traces to measured evidence in this paper (C6). The supplements sharpen attribution within the tested configurations: the historical replay disagreement sits inside its endpoint's same-day nondeterminism floor, replicated over a second five-day round in which one day's own floor stood out (A-S1); the floor appears at comparable magnitude on all four providers measured, unrepaired by any exposed metadata (A-S2); and the readout separates constructed errors by type, not magnitude (A-S3). A cross-track L2 arm closes the ladder from above, on one stack and one load pattern only (\S{}8.3): self-hosted batch-invariant serving beats every shared endpoint while quiet, loses exactness anyway, and falls back into the shared-endpoint range under concurrent load. Whether the shared-endpoint floor originates in batching, kernel scheduling, or deployment rotation is still not identifiable from outside: batching pressure is demonstrated \emph{sufficient} for instability of that magnitude, and no measurement here selects it as the cause.

The proposition we intend to be quoted, scoped to what was measured (the tested providers, their exposed metadata fields, this ranking protocol, these observation windows), is this:

\begin{quote}
\textbf{On a shared endpoint, a model name is not a frozen instrument. A preregistered evaluation must measure its instrument (noise floor, gap distribution, metadata semantics) before freezing any gate on it, because a gate frozen on an unmeasured quantity tests nothing at full ceremony.}
\end{quote}

The constructive path exists at every layer and is priced in this paper: snapshot locking with verification under load (\S{}9.1); operating-characteristic calibration before any freeze, at roughly 2\% of this study's call volume (\S{}9.2); detection-limit conditioning and continuous readouts where ranking is the deliverable (\S{}9.3--9.4); prospective aggregation as the unit of measurement on shared infrastructure (\S{}9.5). None of it requires new access, only the discipline of treating an evaluation instrument as metrology treats any instrument: as the first object of study.

These conclusions are statements about black-box observation of language models through shared serving infrastructure, under this paper's operationalisation; they assert nothing about the internal dynamics of the observed models, for which a separate white-box track with its own preregistration exists. The supplements A-S1/A-S2/A-S3 are complete and reported in full (\S{}10); the checklist (Appendix D) and the evidence packages (\S{}12) are released for reuse.

\bibliography{references}
\bibliographystyle{tmlr}

\appendix
\begin{figure}[t]
  \centering
  \includegraphics[width=\linewidth]{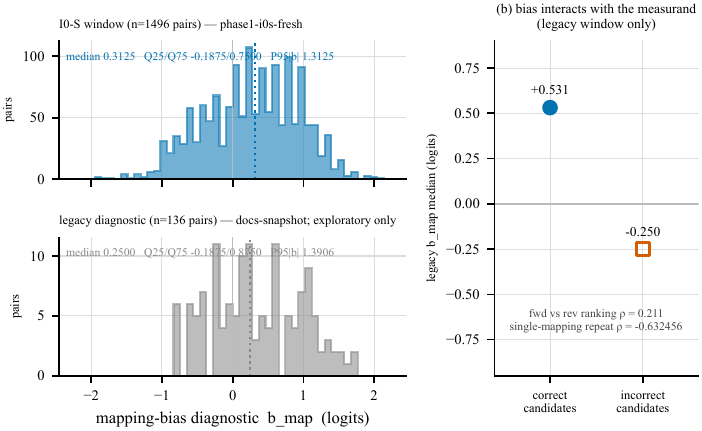}
  \caption{The mapping-bias diagnostic $b_{\mathrm{map}}$ (half the
  difference of the two arms' logits; zero for an unbiased readout) in the
  legacy and fresh I0-S windows, faceted, never pooled. Right: the bias
  interacts with candidate correctness (legacy window; exploratory
  stratified view, no interaction test claimed).}
  \label{fig:bmap}
\end{figure}

\section{Gate definitions, frozen anchors, and the verdict map}

Every confirmatory statement in the main text traces to a gate or reading rule that was frozen before the data it judges existed. This appendix states each one in full: statistic, threshold, denominator, invalidity and tie handling, freeze anchor, and outcome. The verbatim frozen texts (gate documents, preregistrations, sealed configuration snapshots) are members of the archives listed in \S{}12 and ship in the supplement; each entry below names its anchor so the correspondence is checkable file by file.

\subsection{Legacy D1 stability gate (first campaign)}

\emph{Statistic.} Median within-task rank Spearman between repeat readouts under the single-mapping protocol (B.1). \emph{Threshold.} $\geq$ 0.90, frozen before execution. \emph{Denominator.} All repeated task groups with defined Spearman; missing labels invalidate the affected arm. \emph{Outcome.} \textbf{-0.632456} at near-ceiling delivery (1,814/1,816; coverage 99.853\%) \evtag{phase1-oldD1@2dc38b05}; diagnosed in \S{}5.1.

\subsection{Gate I-2R: candidate distinctness (second campaign, Stage 2)}

\emph{Rule.} Within every task, the presented candidates must be pairwise distinct as rendered strings; a round in which any task presents duplicate candidates fails whole. \emph{Outcome.} Round 0 failed (40 of 80 tasks carried pairwise-duplicate candidates); all 160 generations were excluded, and Round 1 reran with fresh identities and tasks (\S{}12.1). Nothing from Round 0 enters any denominator.

\subsection{Gate I-3S: instrument stability (I0-S full run)}

Five subcriteria, all frozen in the sealed gate document \evtag{phase1-i0s-fresh@ac06bb97}:

\begin{center}
\small
\begin{tabularx}{\textwidth}{l>{\raggedright\arraybackslash}Xl>{\raggedright\arraybackslash}Xl}
\toprule
\# & Subcriterion & Frozen threshold & Measured & Verdict \\
\midrule
1 & Single-arm exact two-label coverage & $\geq$ 0.995 & 1.000 & pass \\
2 & Pair completeness & $\geq$ 0.990 & 1.000 & pass \\
3 & Arm label mass, median and 1st percentile & $\geq$ 0.95 / $\geq$ 0.80 & 0.999475 / 0.997172 & pass \\
4 & Repeat scalar stability, median \(|\Delta r_{\mathrm{sym}}|\) & $\leq$ 0.03 & 0.006710 & pass \\
5 & Repeat ranking stability, within-task Spearman median & $\geq$ 0.90 & \textbf{0.400} & \textbf{fail} \\
\bottomrule
\end{tabularx}
\end{center}

\emph{Denominators.} 1,496 executed pairs; the repeat subset is the frozen 10\% re-measurement (136 pairs); subcriterion 5 runs over 32 repeated rank groups, 31 with defined Spearman (one constant on repeat, excluded by the frozen validity rule, never imputed; 29 of 31 below threshold). \emph{Handling.} Missing labels invalidate arm and pair; no reallocation, imputation, or metric substitution; ties receive average ranks (sealed implementation in the supplement).

\subsection{Gate I-5R: next-day exact replay (I0-R, Stage 3-F2)}

\emph{Design.} After a mandatory $\geq$ 24 h gate, the 100 preregistered judge requests are replayed with byte-identical bodies (SHA-256 verified 100/100) against the same model name. \emph{Statistic.} Full-ranking agreement (exact permutation match; "structured-field agreement" in the frozen artifacts). \emph{Threshold.} $\geq$ 0.99. \emph{Denominator.} All 100 replays, canaries kept. \emph{Outcome.} \textbf{78/100 = 0.78} $\rightarrow$ fail \evtag{i0r-round@e8420c2f}. Prospectively frozen inside a campaign not described as fully preregistered (\S{}11.5).

\subsection{A-S1 frozen interpretation bands (descriptive; no pass/fail)}

\emph{Statistic.} \(d = \mathrm{median}(A_{\mathrm{within}}) - \mathrm{median}(A_{\mathrm{between}})\), where \(A\) is full-ranking agreement between window pairs (20 same-day, 25 cross-day pairs from ten windows). \emph{Frozen bands.} \(|d| < 0.03\) reads as a nondeterminism floor; \(d \ge 0.10\) as a systematic cross-day component; intermediate values are reported without a name. \emph{Outcome.} 0.805 - 0.800 = \textbf{0.005} $\rightarrow$ nondeterminism floor (\S{}10.1). Preregistration frozen at commit \texttt{ffbeb95}; run manifests \texttt{069fd364\ldots{}} (Day 1), \texttt{47a78e6a\ldots{}} (Day 2). \emph{Replication round (r2).} Frozen at \texttt{29b5962} with two further reading rules: a replication reading (same bands, five days, separate unit, no pooling) and a per-day spread reading (per-day same-day medians; spread beyond the 0.03 band names the least stable day). Outcomes: \(d = 0.0050\) $\rightarrow$ replicated; spread 0.0450 $\rightarrow$ a day stood out (day 2); five day-manifests \texttt{38a73362\ldots{}}/\texttt{6800952c\ldots{}}/\texttt{ba032eaf\ldots{}}/\texttt{26c25814\ldots{}}/\texttt{35dbe1a2\ldots{}} (\S{}10.1).

\subsection{A-S2 three frozen readings (descriptive; per provider, never merged)}

Preregistration frozen at commit \texttt{cdc012f}; run manifests \texttt{dabb7183\ldots{}} / \texttt{2a37b642\ldots{}}. (i) \emph{Platform-class test}: given valid-readout rate $\geq$ 0.95 on every provider, read \(\max_p \mathrm{median}(A^p_{\mathrm{within}})\) against 0.95; outcome 0.8788 $\rightarrow$ fired. (ii) \emph{Per-provider decomposition}: A-S1's bands per provider; four floors (d = 0.000/0.000/0.000/-0.005). (iii) \emph{Fingerprint predictiveness} at probe-level pairing; outcomes in \S{}8.1.

\subsection{A-S3 reading rules (controlled-error battery)}

Preregistration frozen at commit \texttt{c145f0c}; plan \texttt{487e4e30\ldots{}}, battery \texttt{8f41e2fd\ldots{}}, manifest \texttt{2dd41dd5\ldots{}}. (i) Per-bin resolution: a bin is \emph{resolved} when the 95\% Wilson lower bound of its separation exceeds chance 0.5. (ii) Main verdict, one of four frozen outcomes over the four \(\varepsilon\) bins at or above \(10^{-3}\); realised outcome: \texttt{limit\_below\_smallest\_bin\_tested}. (iii) The within-type contrast governs interpretation over the pooled table (frozen \S{}14.3). (iv) The four-pass replay disagreement and the below-\(10^{-3}\) group are reported but enter no verdict. Outcomes in \S{}10.3.

\subsection{The verdict map}

Every headline verdict, its freeze anchor, and where it is reported:

\begin{center}
\small
\begin{tabularx}{\textwidth}{>{\raggedright\arraybackslash}X>{\raggedright\arraybackslash}X>{\raggedright\arraybackslash}Xl}
\toprule
Verdict & Frozen anchor & Execution anchor & Reported \\
\midrule
Legacy D1 gate fail (-0.632) & sealed gate doc \texttt{2dc38b05\ldots{}} & legacy run directory (\S{}12, \texttt{phase1-oldD1}) & \S{}5.1 \\
I-3S fail (0.400 vs 0.90) & sealed gate doc \texttt{ac06bb97\ldots{}} & run manifest \texttt{207cd6eb\ldots{}} & \S{}4.1, \S{}5.2 \\
I-5R fail (0.78 vs 0.99) & frozen replay plan (\S{}3.4) & run manifest \texttt{22c05608\ldots{}} & \S{}4.1, \S{}5.3 \\
A-S1 nondeterminism floor (d = 0.005) & prereg @ \texttt{ffbeb95} & manifests \texttt{069fd364\ldots{}}, \texttt{47a78e6a\ldots{}} & \S{}10.1 \\
A-S2 platform-class + fingerprint readings & prereg @ \texttt{cdc012f} & manifests \texttt{dabb7183\ldots{}}, \texttt{2a37b642\ldots{}} & \S{}10.2, \S{}8.1 \\
A-S3 limit not bracketed; type-governed separation & prereg @ \texttt{c145f0c} & manifest \texttt{2dd41dd5\ldots{}} & \S{}10.3 \\
L2 arm: quiet 0.9872 / loaded 0.8925 (clean subset); batch invariance necessary, not sufficient & cross-line prereg @ \texttt{33a5f25\ldots{}} (white-box track) & run manifest \texttt{5d34c129\ldots{}}; report hash in the citation ledger & \S{}8.3 \\
\bottomrule
\end{tabularx}
\end{center}

\subsection{Frozen wire configuration by campaign}

The I0-S full run executed under the configuration below, quoted from the authorised snapshot sealed inside the run directory and hash-pinned by its manifest \evtag{phase1-i0s-fresh@75ea5677}:

\begin{center}
\small
\begin{tabularx}{\textwidth}{>{\raggedright\arraybackslash}X>{\raggedright\arraybackslash}X}
\toprule
Item & Value \\
\midrule
Task seed / design & \texttt{20260729}; 80 tasks $\times$ 4 candidates = 320; visible steps 4--8; prefix grid \(J=4\) \\
Generator & \texttt{gpt-4o-mini-2024-07-18} (family \texttt{openai-gpt-4o}), temperature 0.8, strict structured outputs \\
Observer & \texttt{qwen3.7-plus-2026-05-26} (family \texttt{qwen3.7-plus}), shared OpenAI-compatible Chat Completions endpoint (DashScope compatible mode), temperature 0.0, \texttt{max\_completion\_tokens} 1, \texttt{top\_logprobs} 5, thinking disabled \\
Measurement & Protocol \texttt{d1\_symmetric}; mappings \(\{A{\mapsto}+,B{\mapsto}-\}\) and \(\{B{\mapsto}+,A{\mapsto}-\}\); clip \(\varepsilon=10^{-4}\); repeat fraction 0.10; no imputation; no candidate reuse \\
Request plan & 320 generator + 2,720 main observer arms (1,360 pairs) + 272 repeat arms (136 pairs); \textbf{3,312 fixed calls}, hard cap \textbf{3,632}, concurrency 1 \\
Budget & Hard limit CNY 9.99; per-run authorisation row binding models, counts, cap, and budget \\
\bottomrule
\end{tabularx}
\end{center}

The judge-protocol campaigns share one frozen wire body shape, quoted from the sealed request plans (members of the archives in \S{}12): \texttt{temperature} 0.0, \texttt{max\_completion\_tokens} 256, \texttt{response\_format \{"type": "json\_object"\}}, \texttt{seed} 2026080302, thinking disabled via the provider's documented switch. Provider deviations, all frozen before execution and disclosed: \texttt{mistral} rejects \texttt{seed} with HTTP 422, so that field is dropped for that provider only, and its rate-limit headers impose a measured 1.5 s send interval; \texttt{deepseek} disables thinking through \texttt{extra\_body}. For the legacy archives the recorded request hash pins the SDK-parameter shape rather than wire bytes, a limitation stated wherever byte-equivalence is claimed (\S{}11.5).

\subsection{The excluded-data register}

Nothing was deleted, edited, or repaired anywhere in the study (\S{}12.1); every exclusion event, its disposition, and its standing obligation:

\begin{center}
\small
\begin{tabularx}{\textwidth}{>{\raggedright\arraybackslash}X>{\raggedright\arraybackslash}X>{\raggedright\arraybackslash}X}
\toprule
Event & Disposition & Standing obligation \\
\midrule
Stage 2 Round 0: 40 of 80 tasks produced pairwise-duplicate candidates (Gate I-2R fail) & all 160 generations excluded; R1 reran with fresh identities and tasks; an exclusion registry ships with the package & never merged with R1; cost retained \\
Stage 3-E2: wire key-order serialisation defect & all 160 calibration candidates voided; E2-R1 changed key order only & permanently excluded; defect disclosed; CNY 0.141878400 retained \\
Stage 3-B0: 13 calls, all HTTP 403 (workspace endpoint) & retained as raw failure; shared endpoint adopted as an independent replacement window & never cited as capability evidence \\
Phase-1 transport-blocked window: 408 attempts, 0 responses; observer all-403 window: 1,817 attempts & retained as raw failure records & not research diagnostics \\
Stage 3-F1 first freeze: missing local parser/alias-map metadata & voided before live; re-frozen; both manifests retained & first version never used for execution \\
Legacy D1-S diagnostic (136 pairs) & exploratory only & never grounds for declaring I-3S passed \\
Rejected candidate-reuse route (2 dry runs) & fresh restart adopted (legacy provenance unrecoverable, \S{}12.4) & recorded as a rejected design \\
A-S1 v1 Day-1 window: 100 calls, 100 responses, \textbf{0 valid readouts} (an SDK-shape parameter never entered the wire body, so thinking mode stayed enabled and every response hit the token limit) & permanently excluded; r1 re-froze with the wire defect fixed & disclosed as an instrument-failure exhibit; the archived request hash pins SDK-parameter shape, not wire bytes \\
A-S2 r2 Day-1: aborted after 238 attempts (provider rate-limit quota exposed two frozen premises as false) & run left unsealed; never enters any denominator; one in-flight plan ID permanently retired & disclosed; redesign went through full re-freezing, not amendment \\
A-S2 r3 first Day-1: aborted after 618 attempts (a scheduling guard parsed plan IDs by fixed position, never matched A-S2's shape, and silently allowed a 10.6-minute window gap against a frozen $\geq$ 2 h rule) & run left unsealed; excluded; one in-flight plan ID permanently retired; the guard was rewritten to refuse what it cannot parse & disclosed; the frozen schedule itself was correct and unchanged \\
A-S1 r2 first Day-4 (385 attempts): two executor processes concurrently wrote one run directory; window 2 received 285 requests, and 21 probes carry conflicting readings & run left unsealed; voided whole and archived separately (\texttt{a-s1-r2-voided.zip}); day 4 re-ran four days later under a new run group, after a runbook guard against concurrent executors & permanently excluded; cost retained; the round's pre-signed seven-date authorisations (researcher travelling; one date expired unused) are disclosed with it \\
\bottomrule
\end{tabularx}
\end{center}

\subsection{Rule $\leftrightarrow$ evidence map (\S{}9)}

\begin{center}
\small
\begin{tabularx}{\textwidth}{>{\raggedright\arraybackslash}X>{\raggedright\arraybackslash}X>{\raggedright\arraybackslash}X}
\toprule
Rule & Prevented failure & Evidence \\
\midrule
9.1 snapshot locking & replay gates on uncontrolled propositions & \S{}5.3, \S{}8.1--8.3, \S{}10.1--10.2 \\
9.2 pre-freeze OC calibration & unreachable gates at full ceremony & \S{}4.1, \S{}5.2, \S{}7, \S{}10.1 \\
9.3 informative-pair conditioning & detection-limit violations in denominators & \S{}5.2, \S{}10.3 \\
9.4 continuous readouts & combinatorial amplification of the noise floor & \S{}6 \\
9.5 prospective aggregation & single-draw measurements on nondeterministic infrastructure & \S{}5.3, \S{}10.1 \\
9.6 measurand construction & degenerate-by-design quantities & \S{}5.2 \\
9.7 checklist reporting & unauditable instrument status & \S{}3.5, \S{}8, \S{}12 \\
9.8 fail-closed guards & silent default-allow in evaluation tooling & \S{}12.1, A.10 \\
\bottomrule
\end{tabularx}
\end{center}

\subsection{Terminology map}

\begin{center}
\small
\begin{tabularx}{\textwidth}{>{\raggedright\arraybackslash}X>{\raggedright\arraybackslash}Xl}
\toprule
Name & What it is & Home \\
\midrule
D1, D1-S & first campaign's readouts: single-mapping two-label logprob; symmetrised successor & \S{}3.2, \S{}5.1 \\
I0-S & first campaign's full symmetrised run (3,312 calls) & \S{}3.3--\S{}3.4 \\
I0-R & second campaign: structured full-ranking judge & \S{}3.2, \S{}5.3 \\
F2 & I0-R's next-day byte-identical replay window (100 replays) & \S{}5.3 \\
I-3S, I-5R & the two decisive instrument gates & \S{}3.4, A.3--A.4 \\
M1--M3 & failure modes: mapping bias; near-degenerate separations; replay drift & \S{}5 \\
C1--C6 & the paper's claims & \S{}4--\S{}9 \\
A-S1/S2/S3 & supplements: same/cross-day; four providers; constructed errors & \S{}10 \\
L0--L2 & snapshot-identity levels of the auditability ladder & \S{}8.3 \\
\(b_{\mathrm{map}}\), \(z_{\mathrm{sym}}\) & mapping-bias diagnostic; symmetrised readout & \S{}3.2, App. B \\
full-ranking agreement & fraction of replays returning the exact reference permutation & \S{}3.4, \S{}6 \\
\bottomrule
\end{tabularx}
\end{center}

\section{Readout protocol derivation}

\subsection{Two-label logprob readouts (legacy D1 and D1-S)}

The legacy readout instructs the observer to answer with exactly one of two labels and reads both labels' log-probabilities at the single answer-token position: \(r = e^{\ell_+}/(e^{\ell_+}+e^{\ell_-})\), exact (no sampling); a missing label invalidates the arm. Its gate outcome (A.1) showed the readout confounds label-direction preference with the correctness signal (\S{}5.1), motivating the symmetrised successor: each measurement point is asked twice with the label-to-meaning mapping reversed, each arm's probability is clipped at \(\varepsilon = 10^{-4}\) and taken through the logit \(z = \mathrm{logit}(\mathrm{clip}(r, \varepsilon))\), giving

\(z_{\mathrm{sym}} = (z^{(+)} + z^{(-)})/2\), \(\qquad b_{\mathrm{map}} = (z^{(+)} - z^{(-)})/2\).

An unbiased readout has \(b_{\mathrm{map}} = 0\) up to noise; its measured distribution is the M1 evidence (\S{}5.1). Arm order was balanced by frozen seed (748 forward-first, 748 reverse-first); a missing label invalidates arm and pair, with no imputation. The frozen 10\% repeat subset grounds Gate I-3S subcriteria 4--5 (A.3).

\subsection{Structured full-ranking judge (I0-R) and the A-S3 judgment}

The judge returns one JSON object whose \texttt{ranking} field must be an exact permutation of the ten presented aliases; parsing is all-or-nothing, and an invalid readout is retained as a failure, never retried into validity. The readout is the permutation; derived statistics are full-ranking agreement (the frozen decision metric), top-1 agreement, and pairwise agreement (1 minus normalised Kendall distance) (\S{}6). The protocol was selected on frozen development data before the confirmatory stages (\S{}11.5). A-S3 reuses this wire contract on constructed candidates: one \emph{judgment} is the event that the correct candidate strictly outranks a given slip candidate, and a bin's \emph{separation} is the fraction of its judgments that hold, with a 95\% Wilson interval against chance 0.5; bin membership is fixed by the constructed relative error, known before any call (\S{}10.3).

\section{Audit-chain field reference}

Every network attempt appends one pre-call and one post-call record to an append-only JSONL event log inside its run directory. The fields, all recorded verbatim at call time:

\begin{center}
\small
\begin{tabularx}{\textwidth}{>{\raggedright\arraybackslash}X>{\raggedright\arraybackslash}Xl}
\toprule
Field & Content & Checklist item \\
\midrule
request body & the exact JSON sent, stored whole & 16 \\
canonical request SHA-256 & hash of the key-sorted serialisation & 16 \\
wire request SHA-256 & hash of the order-preserving serialisation sent on the wire & 16 \\
raw response & unmodified provider bytes, with SHA-256 & 16 \\
timestamps & UTC, before and after the call & 16 \\
response \texttt{model} & copied verbatim & 1, 2 \\
\texttt{system\_fingerprint} & copied verbatim, including \texttt{null} & 3 \\
usage & prompt/completion token counts as returned & 24 \\
finish reason & copied verbatim & 16 \\
HTTP status & as received & 16 \\
retry count & attempts consumed by this logical request & 16 \\
response headers & credential-shaped values dropped at capture (\S{}12.3) & 16 \\
\bottomrule
\end{tabularx}
\end{center}

Rules around the log, each load-bearing for a claim in the main text: \textbf{reserve-before-network cap} (a counter reserves before each attempt and counts retries and transport failures, so a run cannot exceed its frozen cap; checklist 17); \textbf{strict-prefix resume} (an interrupted run resumes only if the existing log is a byte-exact strict prefix of the frozen plan; checklist 17); \textbf{fail-closed guards} (unparseable input is refused; the one historical fail-open guard is disclosed in A.10 and was rewritten; \S{}9.8, checklist 18); \textbf{sealing} (on completion a manifest records every artifact's SHA-256, the directory becomes immutable, and verification tooling re-hashes every sealed run; checklist 19); \textbf{two-layer reporting} (execution integrity stated separately from scientific verdicts; C1, checklist 19). The implementation ships in the supplement with per-file hashes; the field list can be checked against any public-tier run and the restricted-tier hashes (\S{}12.3).

\section{Reporting checklist for preregistered LLM evaluations}

\emph{Each item is a report-this field, not a best practice; an evaluation that cannot fill an item should say "unmeasured", which is itself a report. Trace: the section whose evidence justifies the item.}

\begin{center}
\small
\begin{tabularx}{\textwidth}{l>{\raggedright\arraybackslash}Xl}
\toprule
\# & Report & Trace \\
\midrule
 & \textbf{D.1 Instrument identity} &  \\
1 & Snapshot-identity level: \textbf{L0} (model name) / \textbf{L1} (pinned snapshot or dedicated deployment) / \textbf{L2} (self-hosted; weights hash, framework, hardware class, determinism settings) & \S{}8.3 \\
2 & Requested model string \textbf{and} the distribution of response-side model strings over the run & \S{}8.4(4) \\
3 & Fingerprint field status \textbf{measured over the run}: absent / churning (distinct-value count) / stable; never taken from documentation & \S{}8.1 \\
4 & For every load-bearing request parameter: the \textbf{probed} behaviour on this provider and model (accepted / rejected / silently ignored) & \S{}8.4(1--2) \\
 & \textbf{D.2 Pre-freeze instrument measurement} &  \\
5 & Noise floor of the readout: repeat-delta distribution from a frozen pilot battery, per aggregation level if \S{}9.5 applies & \S{}5.2 \\
6 & Gap distribution of the measurand: within-group ranges, minimum-gap quantiles, tie rates; where possible a constructed-error pilot \textbf{stratified by both error magnitude and error mechanism/surface form} (a single pooled detection-limit curve can be composition-confounded) & \S{}5.2, \S{}10.3 \\
7 & Fraction of the intended denominator below the declared resolution bound \(\delta\), and the preregistered exclusion rule & \S{}9.3 \\
8 & For platform-stability gates: the same-window agreement floor measured on this endpoint & \S{}10.1 \\
 & \textbf{D.3 Gate design} &  \\
9 & Operating characteristic of every confirmatory gate: pass rates under healthy- and degraded-instrument models, with seed and code & \S{}9.2 \\
10 & The granularity match: which downstream claim consumes the gated statistic, and why metric granularity equals claim granularity & \S{}6.3 \\
11 & For discrete exact-match gates (discouraged): the implied per-comparison stability after combinatorial decay & \S{}6.2 \\
12 & Freeze timestamps for every gate and threshold, and any deviation & \S{}11.5 \\
 & \textbf{D.4 Measurement definition} &  \\
13 & The unit of measurement: single call, or prospective aggregate (\(k\), aggregator, tie rule) fixed before execution & \S{}9.5 \\
14 & Handling of invalid readouts: retained-as-failure vs excluded, fixed in advance; no imputation & \S{}4.2 \\
15 & The prohibition list in force (no replay-selection, no post-hoc metric substitution, no threshold movement) & \S{}4.2, \S{}6.2 \\
 & \textbf{D.5 Execution audit} &  \\
16 & Append-only event log with the Appendix C field set (canonical and wire-order request hashes, raw response hash, UTC timestamps, verbatim metadata, retry counts) & \S{}3.5 \\
17 & Hard request cap (reserve-before-network) and its relation to the frozen plan; resume rule (strict-prefix only) & \S{}3.3, \S{}3.5 \\
18 & Guard behaviour on unparseable input: fail-closed, with evidence the guards were exercised & \S{}9.8 \\
19 & Execution-integrity summary \textbf{separated from} scientific verdicts & \S{}4.3 \\
 & \textbf{D.6 Disclosure and reproduction} &  \\
20 & Excluded-data register: every excluded run with reason, disposition, standing obligation; costs retained & \S{}12.1, A.10 \\
21 & Request-attempt ledger by scope, with the non-pooling declaration (audit volume $\neq$ sample size) & \S{}12.2 \\
22 & Reproduction package tiering: public contents; restricted layer listed \textbf{by hash} & \S{}12.3 \\
23 & Provenance status of every cited run: sealed in-run vs external-pointer (capped at exploratory) & \S{}12.4 \\
24 & Cost ledger with the authoritative-source declaration, including excluded and aborted runs & \S{}12.5 \\
\bottomrule
\end{tabularx}
\end{center}

\emph{Adoption note.} Items 1--4 and 16--19 are collectible by instrumentation alone; 5--11 need a pilot battery (roughly 2\% of this study's call volume; \S{}9.2); 12--15 and 20--24 are writing discipline. Nothing requires access beyond what the evaluated API already returns.

% The full run-directory index (appendix E of earlier drafts) ships in the
% supplement as the store's own searchable index; see section 12.3.

\end{document}